\documentclass[10pt,twocolumn,letterpaper]{article}

\usepackage{iccv}              

\usepackage{amsmath}
\usepackage{color}
\usepackage[dvipsnames]{xcolor}
\usepackage{tikz}
\usetikzlibrary{decorations.text, decorations.pathreplacing, positioning, shapes.geometric, calc}
\usepackage{adjustbox}
\usepackage{multirow}
\usepackage{makecell}
\usepackage{tkz-kiviat,numprint,fullpage} 
\usepackage{pgfplotstable} 
\usepackage{graphicx}
\usepackage{amsmath}
\usepackage{amssymb}
\usepackage{booktabs}
\usepackage{adjustbox}
\usepackage{xtab}
\usetikzlibrary{arrows}
\usepackage[table]{xcolor}  

\usepackage{pifont}
\newcommand{\cmark}{\ding{51}}

\pgfplotsset{compat=1.18}
\usepackage[perpage,symbol]{footmisc}
\definecolor{iccvblue}{rgb}{0.21,0.49,0.74}

\usepackage[pagebackref,breaklinks,colorlinks,allcolors=iccvblue]{hyperref}

\def\paperID{14128} 
\def\confName{ICCV}
\def\confYear{2025}

\title{MultiADS: Defect-aware Supervision for Multi-type Anomaly Detection and Segmentation in Zero-Shot Learning}

\author{Ylli Sadikaj\textsuperscript{*} \textsuperscript{\rm 1 \rm 5} \textsuperscript{\S},
Hongkuan Zhou\textsuperscript{*} \textsuperscript{\rm 2 \rm 3},
Lavdim Halilaj\textsuperscript{\rm 2},\\
Stefan Schmid\textsuperscript{\rm 2},
Steffen Staab\textsuperscript{\rm 3 \rm 4},
Claudia Plant\textsuperscript{\rm 1 \rm 6}\\\\
\textsuperscript{\rm 1}Faculty of Computer Science, University of Vienna, Vienna, Austria\\
\textsuperscript{\rm 2}Bosch Corporate Research, Robert Bosch GmbH, Renningen, Germany\\
\textsuperscript{\rm 3}University of Stuttgart, Stuttgart, Germany, 
\textsuperscript{\rm 4}University of Southampton, Southampton, UK\\
\textsuperscript{\rm 5}UniVie Doctoral School Computer Science, University of Vienna, Vienna, Austria\\
\textsuperscript{\rm 6}ds:UniVie, Vienna, Austria\\
{\tt\small \{ylli.sadikaj, claudia.plant\}@univie.ac.at, steffen.staab@ki.uni-stuttgart.de}\\
{\tt\small \{hongkuan.zhou, lavdim.halilaj, stefan.schmid5\}@de.bosch.com}\\
}

\begin{document}
\maketitle
\definecolor{network-blue}{RGB}{165, 192, 221}
\definecolor{light-yellow}{RGB}{238, 233, 218}
\definecolor{light-green}{RGB}{129, 184, 113}
\definecolor{light-red}{RGB}{242, 182, 160}
\definecolor{light-blue}{RGB}{124, 150, 171}
\definecolor{light-orange}{RGB}{255,147,0}
\definecolor{light-purple}{RGB}{150,115,166}
\definecolor{dark-green}{RGB}{85, 124, 86}
\definecolor{dark-red}{RGB}{217, 22, 86}
\definecolor{purple}{RGB}{155, 126, 189}
\definecolor{dark-purple}{RGB}{59, 30, 84}

\begin{abstract}
Precise optical inspection in industrial applications is crucial for minimizing scrap rates and reducing the associated costs.
Besides merely detecting if a product is anomalous or not, it is crucial to know the distinct types of defects, such as a bent, cut, or scratch. 
The ability to recognize the ``exact" defect type enables automated treatments of the anomalies in modern production lines. 
Current methods are limited to solely detecting whether a product is defective or not, without providing any insights into the defect type, but nevertheless detecting and identifying multiple defects. 
We propose MultiADS, a zero-shot learning approach, able to perform \textbf{Multi}-type \textbf{A}nomaly \textbf{D}etection and \textbf{S}egmentation.
The architecture of MultiADS comprises CLIP and extra linear layers to align the visual and textual representation in a joint feature space. 
To the best of our knowledge, our proposal is the first approach to perform a multi-type anomaly segmentation task in zero-shot learning.
Contrary to the other baselines, our approach i) generates specific anomaly masks for each distinct defect type, ii) learns to distinguish defect types, and iii) simultaneously identifies multiple defect types present in an anomalous product. 
Additionally, our approach outperforms zero/few-shot learning SoTA methods on image-level and pixel-level anomaly detection and segmentation tasks on five commonly used datasets: MVTec-AD, Visa, MPDD, MAD, and Real-IAD. 
\end{abstract}

\vspace{-1em}
\section{Introduction}
\label{sec:intro}
\input{tikz/introduction_figure_v2}
\footnotetext[1]{Both authors contributed equally to this work.}
\footnotetext[4]{Work done during PhD Sabbatical at Bosch Corporate Research.}
One of the primary objectives of the manufacturing industries is to utilize their assembly lines for a wide range of product types. 
Modern factories are equipped with sophisticated and adaptable mechanisms allowing for a quick reconfiguration to various scenarios~\cite{math11030601}.
By doing so, the probability of outputting defective products is significantly increased.
Therefore, to achieve intelligent manufacturing and prevent downtimes, rework, or quality losses, it is essential to detect anomalies promptly and with high precision~\cite{RUEDIGERFLORE2024138, 8997079}.
More concretely, identifying the specific defect\footnote{We use \textit{defect} and \textit{anomaly} terms interchangeably.} type in a product helps operators to understand the underlying causes and effectively implement preventive measures. 
In this regard, optical inspection via visual anomaly detection and segmentation is crucial to identify abnormal products and locate anomalous regions.

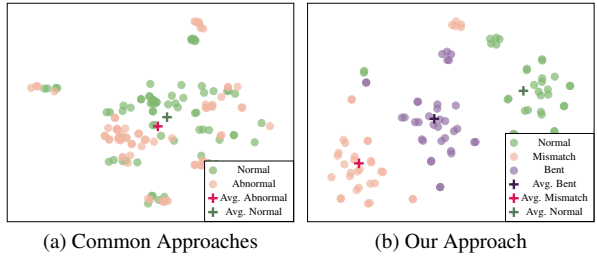
\begin{figure}[t]
\begin{center}
    \begin{subfigure}[b]{.23\textwidth}
    \begin{adjustbox}{width=\textwidth}
    \begin{tikzpicture}
        \begin{axis}[
            grid=minor,
            width=10cm,
            height=8cm,
            ticks=none,
            legend style={
            at={(1,0)}, 
            anchor=south east, 
            font=\small 
    }
        ]
        
        \addplot[
            only marks,
            mark=*,
            mark size=3pt,
            light-green,
            opacity=0.7
        ]
        table[x=x, y=y, col sep=comma] {data/normal.csv};
        \addlegendentry{Normal}
        \addplot[
            only marks,
            mark=*,
            mark size=3pt,
            light-red,
            opacity=0.7
        ]
        table[x=x, y=y, col sep=comma] {data/abnormal.csv};
        \addlegendentry{Abnormal}
        
        \addplot[
            only marks,
            mark=+,
            dark-red,   
            mark size=4pt,
            mark options={line width=2pt}
        ]
        coordinates {(-3.74, -6.48)};
        \addlegendentry{Avg. Abnormal}

        \addplot[
            only marks,
            mark=+,
            dark-green,   
            mark size=4pt,
            mark options={line width=2pt}
        ]
        coordinates {(-2.08,-1.81)};
        \addlegendentry{Avg. Normal}
        
        \end{axis}
    \end{tikzpicture}
    \end{adjustbox}
    \caption{Common Approaches}\label{fig:commonApproaches}
    \end{subfigure}%
    \hfill
    \begin{subfigure}[b]{.23\textwidth}
    \begin{adjustbox}{width=\textwidth}
    \begin{tikzpicture}
        \begin{axis}[
            grid=minor,
            width=10cm,
            height=8cm,
            ticks=none,
            legend style={
            at={(1,0)}, 
            anchor=south east, 
            font=\small 
            }
        ]
        
        \addplot[
            only marks,
            mark=*,
            mark size=3pt,
            light-green,
            opacity=0.7
        ]
        table[x=x, y=y, col sep=comma] {data/normal1.csv};
        \addlegendentry{Normal}
        \addplot[
            only marks,
            mark=*,
            mark size=3pt,
            light-red,
            opacity=0.7
        ]
        table[x=x, y=y, col sep=comma] {data/scratch.csv};
        \addlegendentry{Mismatch}

        \addplot[
            only marks,
            mark=*,
            mark size=3pt,
            light-purple,
            opacity=0.7
        ]
        table[x=x, y=y, col sep=comma] {data/defective_painting.csv};
        \addlegendentry{Bent}
        
        \addplot[
            only marks,
            mark=+,
            dark-purple,   
            mark size=4pt,
            mark options={line width=2pt}
        ]
        coordinates {(-1.45, 0.71)};
        \addlegendentry{Avg. Bent}

        \addplot[
            only marks,
            mark=+,
            dark-red,   
            mark size=4pt,
            mark options={line width=2pt}
        ]
        coordinates {(-6.6,-2.63)};
        \addlegendentry{Avg. Mismatch}

        \addplot[
            only marks,
            mark=+,
            dark-green,   
            mark size=4pt,
            mark options={line width=2pt}
        ]
        coordinates {(4.64,2.81)};
        \addlegendentry{Avg. Normal}
        
        \end{axis}
        
    \end{tikzpicture}
    \end{adjustbox}
    \caption{Our Approach}\label{fig:ourApproach}
    \end{subfigure}
     \vspace{-0.1cm}
    \caption{\small{Visualization of text prompts (TP) embeddings of common approaches and ours for Bracket Brown product of the MPDD dataset utilizing visualization tool t-SNE~\cite{tsne}. Dot signs ($\cdot)$ represent TP embeddings, plus signs ($+$) represent the average embedding of TPs with the same color.} 
    }
    \vspace{-0.9cm}
    \label{fig:latent_space}
\end{center}
\end{figure}

Recent approaches utilize prior knowledge in pre-trained models like CLIP~\cite{CLIP} or DINO~\cite{DBLP:conf/iccv/CaronTMJMBJ21} to boost the generalization performance across a wide range of products for anomaly detection. 
CLIP-based approaches, such as~\cite{WinCLIP, April-GAN, AnomalyCLIP}, employ CLIP knowledge and adapt it for anomaly detection and segmentation by defining text-prompts for normal and abnormal states (cf. Figure~\ref{fig: motivation}a).
Next, they compare the similarity between the image embedding and the average text embedding from generic sets of good and bad prompts.
Thus, they are not exploiting anomaly-relevant knowledge, such as defect types, embedded in pre-trained vision language models (VLMs). 
On the other hand, fine-tuning in the specific domain often leads to overfitting on the training dataset~\cite{zhou2025robustvisualrepresentationlearning}, causing the model to lose valuable knowledge critical for accurate anomaly detection and segmentation. 
In Figure~\ref{fig:commonApproaches}, we visualize how averaging normal and abnormal text embeddings can lead to significant information loss. 


In this paper, we present MultiADS, a zero-shot learning approach for multi-type anomaly detection and segmentation that leverages the prior knowledge of the common defect types in VLMs. 
It aligns the image embedding and the mean text embedding from a general set of good prompts and defect-specific sets of bad prompts. 
As illustrated in Figure~\ref{fig: motivation}b, through our approach, we can answer correctly all three questions, including the question regarding the defect type.
Figure~\ref{fig:ourApproach} shows that MultiADS preserves the meaningful semantic representation within the latent space and clearly distinguishes normal state and distinct defect types. 
Contrarily, competitive baselines could fail to separate between normal and abnormal states, as shown in Figure~\ref{fig:commonApproaches}.
We conduct experiments on five datasets for anomaly detection and anomaly classification, MVTec~\cite{MVTec}, VisA~\cite{VisA}, MPDD~\cite{MPDD}, MAD (real and simulated)~\cite{PAD}, and Real-IAD~\cite{realIAD}. We conducted evaluations in both zero-shot/few-shot settings. 
The empirical results demonstrate that incorporating defect-type information into the learning pipeline improves anomaly detection and segmentation performance across these five datasets. 
We summarize the key contributions as follows: 
\begin{itemize}
    \item Our MultiADS detects multiple defects of the same and/or different types in an anomalous product. 
    Thus, we propose a new task, namely a multi-type anomaly detection and segmentation task, that aims to determine the defect type at the pixel level. We position MultiADS as a baseline in such a new task.
    \item We show that by leveraging anomaly-specific knowledge in pre-trained VLMs, MultiADS further improves its detection and segmentation performance.
    \item We present a Knowledge Base for Anomalies (KBA), that enhances the description of defect types. It can be utilized for defect-aware text prompt construction and facilitates the fine-tuning process of VLMs for anomaly detection and segmentation.        
    \item Additionally, we evaluate the performance of MultiADS on anomaly detection and segmentation against 12 baselines both zero-shot/few-shot settings. The code implementation is publicly available at: \url{https://github.com/boschresearch/MultiADS}.
\end{itemize}

\section{Related Work}
\label{sec:relatedwork}

In this section, we review the most relevant literature based on their learning paradigms and highlight how our approach distinguishes itself from existing methods.

\textbf{Unsupervised Anomaly Detection.}  
There exists a wide variation in the characteristics of objects and their defects, including differences in color, texture, size, and shape. This heterogeneity leads to an extensive range of defect types, making it challenging to compile a representative set of anomaly samples for training data. 
Thus, unsupervised anomaly detection approaches, such as~\cite{bergmann2022beyond, xie2023, Roth_2022, huang2022}, require only normal images for training. These methods typically model images without anomalies and classify any deviations from the learned representation as anomalies. 

\textbf{Zero-Shot Anomaly Detection (ZSAD).} Recent studies have leveraged the power of large-scale VLMs such as CLIP~\cite{CLIP} to perform anomaly detection without any target-specific training.
The success of prompt learning in natural language processing has inspired methods such as CoOp~\cite{CoOp} and CoCoOP~\cite{CoCoOp}, which automatically learn task-specific prompt contexts from only a few labeled examples.
Early methods such as WinCLIP~\cite{WinCLIP} and April-GAN~\cite{April-GAN} adapt CLIP by designing text prompts that differentiate “normal” from “abnormal” states. Also, they introduce window-based strategies or additional linear layers to enhance image segmentation performance.

Other approaches apply the same differentiation technique while adapting the construction for text prompt states. Thus, AnomalyCLIP~\cite{AnomalyCLIP} learns object-agnostic text prompts to capture generic cues of abnormality, 
SimCLIP~\cite{simclip} further adopts implicit prompt tuning.  Similarly, FiLo~\cite{filo} and AdaCLIP~\cite{AdaCLIP} enhance localization by replacing generic anomaly descriptions with adaptively learned fine-grained prompts or tuning hybrid learnable prompts by combining static and dynamic prompts. Contrary to other models, ClipSAM~\cite{clipsam} proposes a novel collaboration between CLIP and SAM~\cite{sam}, whereas MuSc~\cite{MuSc} detects anomalies by exploiting mutual scoring across unlabeled test images.

\textbf{Few-Shot Anomaly Detection (FSAD).} 
FSAD models, such as~\cite{TDG, RegAD, PatchCore, DifferNet}, include several normal sample images from the target domain to train their model.
PromptAD~\cite{PromptAD} refines the image–text alignment process by concatenating normal prompts with anomaly-specific suffixes. GraphCore~\cite{GraphCore} employs graph neural networks to capture rotation-invariant features from limited normal samples, while KAGprompt~\cite{KAGprompt} constructs a kernel-aware hierarchical graph among multi-layer visual features. 
Other methods adopt reconstruction or feature-matching strategies—such as FastRecon~\cite{FastRecon} and FOCT~\cite{foct}-to reconstruct normal appearances from a limited set of normal samples. 
Given the scarcity of anomalous samples, Anomalydiffusion~\cite{anomalydiffusion} proposes to employ a latent diffusion model along with spatial anomaly embeddings to generate authentic anomaly image–mask pairs. 
Meanwhile, AnomalyGPT~\cite{AnomalyGPT} is an interactive method integrating VLMs to provide defect-specific descriptions for a context-aware inspection. 
AnomalyDINO~\cite{AnomalyDINO} uses DINOv2~\cite{DINOv2} to extract robust patch-level features for FSAD.


A major limitation of existing vision-language ZSAD and FSAD methods is their binary focus—only distinguishing between normal and abnormal states, as illustrated in Figures~\ref{fig: motivation} and \ref{fig:latent_space}. 
In contrast, MultiADS is designed to perform multi-type anomaly segmentation by constructing defect-specific text prompts that capture rich semantic attributes. This allows MultiADS to not only detect whether an image is anomalous but also to segment and classify the specific type of defect present - a capability that is critical for automated optical inspection in industrial applications.

\section{Preliminaries}
Here, we introduce the preliminary definitions of binary and multi-type anomaly detection and segmentation, as well as the backbone model. 

\subsection{Binary Detection and Segmentation}
Let $\mathcal{D}_{\text{train}}$ and 
$\mathcal{D}_{\text{target}}$ 
denote two different datasets, training and target datasets, respectively.
Both datasets consist of $X, Y$, where $X = \{\mathbf{x}_i\}_{i=1}^N$ with $N$ images, and $Y = \{(\mathbf{M}_i, y_i)\}_{i=1}^{N}$ with ground truth labels. Each image $\mathbf{x}_i \in \mathbb{R}^{H \times W}$ is masked with $\mathbf{M}_i$ and labeled with $y_i$, where $y_i \in \{0,1\}$ is the indicator for anomaly or not and $\mathbf{M}_i \in \{0,1\}^{H \times W}$ represents the binary anomaly map. Binary anomaly detection and segmentation (BADS) aim to determine if the given image $\mathbf{x}$ contains anomalies and also locate regions in an image that contain anomalies.


\subsection{Multi-type Anomaly Segmentation}
$\mathcal{D}_{\text{train}}$ and 
$\mathcal{D}_{\text{target}}$ 
denote the training and target datasets, respectively.
Both datasets consist of $X$, $Y^\prime$, where $X = \{\mathbf{x}_i\}_{i=1}^N$ with $N$ images and  $Y^{\prime} = \{\mathbf{M}^{\prime}_i\}_{i=1}^{N}$. Each image $\mathbf{x}_i$ is labeled with $\mathbf{M}^{\prime}_i\in \{0,1,..,K\}^{H \times W}$, representing the multi-defect segmentation map for one normal class and $K$ abnormal classes. Multi-type anomaly segmentation (MTAS) aims to locate the anomalies and identify various anomaly types.

\subsection{Backbone Model}
Contrastive Language Image Pre-training (CLIP) is a large-scale vision-language model pre-trained on million-scale image-text pairs, $\{(x_i, t_i)\}^{N}_{i=1}$.
It encompasses an image feature encoder, $f(\cdot)$, and a text feature encoder, $g(\cdot)$. 
CLIP aims to maximize the correlation between $f(x_i)$ and $g(t_i)$ utilizing cosine similarity. 
Thus, for a given image input $x$ and a closed set of text $T = \{t_1, \dots, t_K\}$, representing the text prompt for $K$ classes, CLIP performs classification as follows:
\begin{equation}
    p(y = j|x) := \frac{exp(\langle f(x), g(t_j)\rangle/\tau)}{\sum_{i = 1}^{K}{exp(\langle f(x), g(t_i)\rangle/\tau)}} \text{,}
    \label{eq:cosSimilarity}
\end{equation}
where $\tau > 0$ is the temperature hyperparameter, whereas $\langle \cdot, \cdot \rangle$ represents the cosine similarity. 

\section{MultiADS Approach}
\input{tikz/architecture}
Our proposed approach is a CLIP-based model adapted for zero-shot and few-shot learning for detecting anomalies and identifying the defect types in images from the manufacturing domain. 
It learns the alignment of image features with their corresponding text features that represent a distinct defect type, as shown in Figures~\ref{fig: motivation} and \ref{fig:architecture}. Anomaly maps constructed for each distinct defect type enable multi-class defect detection and segmentation.  



\textbf{Knowledge Base for Anomalies.} 
We leverage the meta-data from established industrial defect detection datasets, including MVTec-AD, VisA, MPDD, MAD (real and simulated), and Real-IAD, to acquire comprehensive defect-aware information for each product class. 
Additionally, we incorporate supplementary defect-type properties (attributes) into our knowledge base for anomalies (KBA), including size and shape. 

Initially, we group the defect types into superclasses, such that \textit{bent}, \textit{bent lead}, and \textit{bent wire} are represented by the \textit{bent} superclass, similarly \textit{scratch}, \textit{scratch head}, and \textit{scratch neck} are under \textit{scratch}. 
Thus, we have abstract classes like \textit{bent}, \textit{cut}, \textit{scratch}, capturing all possible defect types that can occur in a given dataset. 
Details of the acquired information for all datasets part of our KBA are given in the Appendix.

\textbf{Defect-aware Text Prompts.} 
Next, we utilize the constructed KBA as prior knowledge for our text-prompt construction, as illustrated in Figures~\ref{fig: motivation}b and Figure~\ref{fig:architecture}. 
We select the same set of variations of text samples as in~\cite{WinCLIP, April-GAN} to construct text prompts for each given defect class.
Figure~\ref{fig:latent_space} shows the difference between other baselines and our approach regarding the text prompt embeddings.
More details for defect-aware text prompts are provided in the Appendix.

\subsection{Training Phase}
An overview of the training phase of our proposed method is shown in Figure~\ref{fig:architecture} (LHS). 
We use different datasets for training and testing with their respective prompt set numbers denoted by $K_1$ and $K_2$. 

\subsubsection{Image and Text Embedding}
Each image $\mathbf{x}$ is provided as input to the image encoder to get image patch embeddings at $m$ different stages during encoding, as in~\cite{April-GAN, AnomalyCLIP}, $\mathbf{E}^p_i \in \mathbb{R}^{h\times w \times N_i}, i\in\{0,1,...,m\}$ with the resolution $h \times w$ and layer $N_i$, as well as one global image embedding $\mathbf{z}^x \in \mathbb{R}^{N_z}$. 
We use $K_1+1$ sets of text prompts: one representing the normal state and $K_1$ representing abnormal states corresponding to $K_1$ defect types. 
Each set of text prompts is fed into the CLIP text encoder, and we obtain an averaged text embedding for each set by averaging the embeddings of individual prompts. 
This process yields $K_1+1$ averaged text embeddings $\mathbf{z}^t \in \mathbb{R}^{N_z}$, each representing a distinct state.

\subsubsection{Aligning Image Patches and Text Prompts}

The visual encoder of CLIP is originally trained to align the global object embeddings with text embeddings.  
To align the two embedding spaces, visual - extracted by the CLIP image encoder, and textual  - extracted by the CLIP text encoder, we utilize adapters consisting of a single linear learnable layer. 
For image patch embeddings at each stage $i$, a linear adapter takes $\mathbf{E}^p_i$ as input and outputs $\mathbf{Z}^p_i \in \mathbb{R}^{h\times w \times N_z}$. They are compared with $K_1+1$ text embeddings $\mathbf{z}^t$ to get the similarity map. 
Since we choose image patches embeddings at $m$ different stages, we get $m$ similarity maps $\mathbf{S}_i \in \mathbb{R}^{(K_1+1) \times h \times w}$, where $h, w$ are the resolution of the similarity maps, $K_1$ is the number of defect types. 
Each map $\mathbf{S}_i$ is up-sampled to match the size of the input image and aligned with the ground truth segmentation map $\mathbf{M}^{\prime}_x$. 

\subsubsection{Training Objective}


Two typical losses, focal~\cite{focalloss} and dice~\cite{diceloss}, are used for segmentation tasks. 
Focal loss is designed to address class imbalance issues, especially in tasks like object detection, where there is often a significant imbalance between classes.
We face the same challenge, i.e., a high number of normal images and a low number of abnormal images; therefore, we apply a multi-class focal loss for multi-defect segmentation
along with the binary dice loss for anomaly segmentation. 
These two training objectives are combined to form the final loss function: 

\vspace{-1.5em}
\begin{multline}
    \mathcal{L} = \sum_{i=1}^{m} \mathcal{L}_\text{focal}(UP(\mathbf{S}_i), \mathbf{M}^{\prime}_x) + \\ \mathcal{L}_\text{dice}(\mathbf{1} - UP(\mathbf{S}_i)[0], \mathbf{M}_x) \text{,}
\end{multline}


where \( \mathbf{M}^{{\prime}}_x \) represents the ground truth multi-defect segmentation map, and \( \mathbf{M}_x \) is the binary anomaly map.  $UP(\cdot)$  denotes the up-sampling function used to scale the similarity map to the input image resolution. Note that in the training phase, the global anomaly score $a_x$ is not fine-tuned. 

\subsection{Inference Phase}
To test the trained model's performance in the target dataset, we first construct $K_2+1$ sets of text prompts, representing one normal state without defect and $K_2$ states representing distinct defect types of the target domain. 
An overview of the inference phase of our proposed method is shown in Figure~\ref{fig:architecture} (RHS).

Each set of text prompts is input into the CLIP text encoder to generate embeddings, while the query image is passed through the CLIP image encoder and then the adapter to produce $m$ similarity maps $\mathbf{S}_i \in\mathbb{R}^{ (K_2+1) \times h \times w}$.
The respective similarity maps are then up-sampled to match the original size of the input image. 
The multi-defect segmentation map is calculated by averaging the up-sampled similarity map:

\begin{equation}
    \mathbf{\hat{M}}_x^{\prime} = \frac{1}{m}\sum_{i=1}^{m} UP(\mathbf{S}_i)\text{.}
\end{equation}
We only take the \textit{first} layer of similarity maps and perform a complement operation on each pixel to create the anomaly score map. Since there are $m$ similarity maps, we average the $m$ anomaly score maps to obtain the final anomaly map:
\begin{equation}
    \mathbf{\hat{M}}_x = \frac{1}{m}\sum_{i=1}^{m}\mathbf{1} - UP(\mathbf{S}_i)[0] \text{.}
\end{equation}

The global image embedding $\mathbf{z}^x$ from the pre-trained CLIP image encoder is also compared with $K_2+1$ text embeddings to get $K_2+1$ global similarity scores. 
After the normalization, the complement of the similarity score compared to the normal state text prompts is used as the final global anomaly score $a_x$. 
We perform zero-shot learning based on the acquired anomaly map $\mathbf{\hat{M}}_x$ and global anomaly score ${a}_x$. 
Few-shot learning is conducted based on the acquired anomaly map $\mathbf{\hat{M}}_x$, global anomaly score $a_x$, and reference anomaly map $\mathbf{\hat{M}}_{\text{ref}}$ between query image and reference normal image(s). 

\subsubsection{Multi-type Anomaly Segmentation}
The $m$ similarity maps $\boldsymbol{S}_i, i \in \{1,\dots,m\}$, are up-sampled to match the input image size and then averaged to produce the multi-defect segmentation map, \( \mathbf{\hat{M}}_x^{\prime} \in \mathbb{R}^{(K_2+1) \times h \times w} \). This map captures both the anomaly locations and their respective defect types, enabling effective support for the multi-type anomaly segmentation task.

\subsubsection{Zero-shot Learning}
For zero-shot learning, the output anomaly map $\mathbf{\hat{M}}_x$ is used for anomaly segmentation and compared with the ground truth labels. 
The highest anomaly score: $\max{(\mathbf{\hat{M}}_x)}$ on anomaly map and global anomaly score $a_x$ are averaged and then compared against a threshold $\theta$ to determine whether the image contains an anomaly.


\subsubsection{Few-shot Learning}
To conduct few-shot learning, we need to compute an extra reference anomaly map based on the similarity between the query image and several reference normal images.
The reference normal image(s) are fed into the image encoder to get $m$ stages of image patch embeddings.
We leverage memory banks~\cite{April-GAN} to store the features of the reference images, which can be compared with input image features by cosine similarity to obtain the reference anomaly map $\hat{M}_\text{ref}$. The final anomaly map $\mathbf{\hat{M}}_{\text{final}} = \frac{1}{2}(\mathbf{\hat{M}}_x + \mathbf{\hat{M}}_{\text{ref}})$ is used for anomaly segmentation. $\mathbf{\hat{M}}_{\text{final}}$ instead of $\mathbf{\hat{M}}_x$ is used to determine the anomaly itself.

\subsubsection{Filtering Out Product-irrelevant Defect Types}
For a specific product type, only certain defect types are relevant. 
During the inference phase, this filtering step involves excluding text prompt sets associated with defect types that are not applicable to the product, ensuring that only relevant defect types are considered. 
Here, the method that includes this filtering process is referred to as MultiADS-F, while the original version without filtering remains as MultiADS.


\section{Experiments}

In this section, we describe datasets and baselines and discuss the results of the conducted experiments. 

\subsection{Datasets}
Five common datasets: MVTec-AD~\cite{MVTec}, VisA~\cite{VisA}, MPDD~\cite{MPDD},  MAD (simulated and real)~\cite{PAD}, and Real-IAD~\cite{realIAD} are used for the multi-type anomaly segmentation as well as the binary anomaly detection and segmentation task, respectively. 
More details of these datasets are provided in the Appendix.

\subsection{Experiment Setting}
We adopt a transfer learning setting, where the model is trained on one of the datasets and evaluated on the remaining. In the zero-shot learning scenario, the trained model is directly applied to the target dataset without any additional information from the target dataset. In contrast, the few-shot learning scenario allows the trained model to access a small number of normal images from the target dataset for further adaptation. 

We use the ViT-L-14-336 CLIP backbone from OpenCLIP~\cite{ilharco_gabriel_2021_5143773}, pre-trained on the LAION-400M\_E32 setting of open-clip. The learning rate is set to 0.001, with a batch size of 8. The stage number $m=4$. The features are selected from layers: 6, 12, 18, and 24. 

\subsection{Evaluation Metrics}
We assess the anomaly detection performance on zero/few-shot learning settings with three metrics, namely the receiver-operator curve (AUROC), the F1-score at the optimal threshold (F1-max), and the average precision (AP). 
Similar to ~\cite{WinCLIP, AnomalyCLIP, April-GAN}, the anomaly segmentation is quantified by AUROC, F1-max, and the per-region overlap (PRO) of the segmentation using the pixel-wise anomaly scores.
For the multi-type anomaly segmentation task, we employ AUROC, F1-score, and AP with the macro averaging setting. 

\subsection{Baselines}
We compare the performance of our approach with the following 12 baselines:  CLIP~\cite{CLIP}, CLIP-AC~\cite{CLIP}, CoOp~\cite{CoOp}, CoCoOp~\cite{CoCoOp}, PatchCore~\cite{PatchCore}, WinCLIP~\cite{WinCLIP}, April-GAN~\cite{April-GAN}, InCTRL~\cite{InCTRL}, PromptAD~\cite{PromptAD}, AnomalyCLIP~\cite{AnomalyCLIP}, AdaCLIP~\cite{AdaCLIP}, and AnomalyGPT~\cite{AnomalyGPT}. 
CLIP, CLIP-AC, CoCo, CoCoOP, WinCLIP, April-GAN, AnomalyCLIP, and AdaCLIP are zero-shot learning approaches. Whereas CoOp, WinCLIP, and April-GAN can also learn in the few-shot setting, as other approaches, PatchCore, PromptAD, InCTRL, and AnomalyGPT. The comparison of batch zero-shot setting with MuSc~\cite{MuSc} and AnomalyDINO~\cite{AnomalyDINO} is discussed in the Appendix. We did not include other baselines such as~\cite{filo, clipsam, simclip} because their authors did not provide implementation yet. 


In the evaluation process, we use the basic approach, MultiADS, and the filtering-based variant, MultiADS-F.

\subsection{Results}

Next, we present and discuss results from the experiments for multi-type anomaly segmentation in zero-shot settings and binary ZSAD and FSAD.

\subsubsection{Multi-type Anomaly Segmentation}

First, we discuss our MultiADS's performance in the new task, the multi-type anomaly segmentation (MTAS) task, which can segment various defect types. 
To the best of our knowledge, we are the first to perform such a task, and thus we present MultiADS as a baseline.

Table~\ref{tab:multi-defect} shows the results of MultiADS on the MTAS task in a zero-shot learning setting.  
We observe that our approach achieves high accuracy in terms of the AUROC metric for pixel-level segmentation of distinct defects in all datasets. As expected, MultiADS performs with higher accuracy in terms of AP metric on datasets with fewer anomaly types, such as MPDD and MAD-real, and the accuracy is slightly lower on datasets with multiple anomaly types appearing concurrently, such as Real-IAD and VisA. Additionally, we found that MultiADS performs slightly better on the VisA dataset when our model is trained on the MVTec-AD or Real-IAD datasets rather than the MPDD dataset due to higher similarity between defect types of the VisA dataset with MVTec-AD and Real-IAD datasets. Similarly, the VisA dataset serves as a good model trainer regarding the performance of the model on the MVTec-AD dataset.
In summary, these results indicate that MultiADS can successfully differentiate between various defect types. We provide more results on the MTAS task in the Appendix.

\begin{figure}[ht]
    \centering
    \begin{subfigure}[b]{0.22\textwidth}
        \centering
        \includegraphics[width=\textwidth]{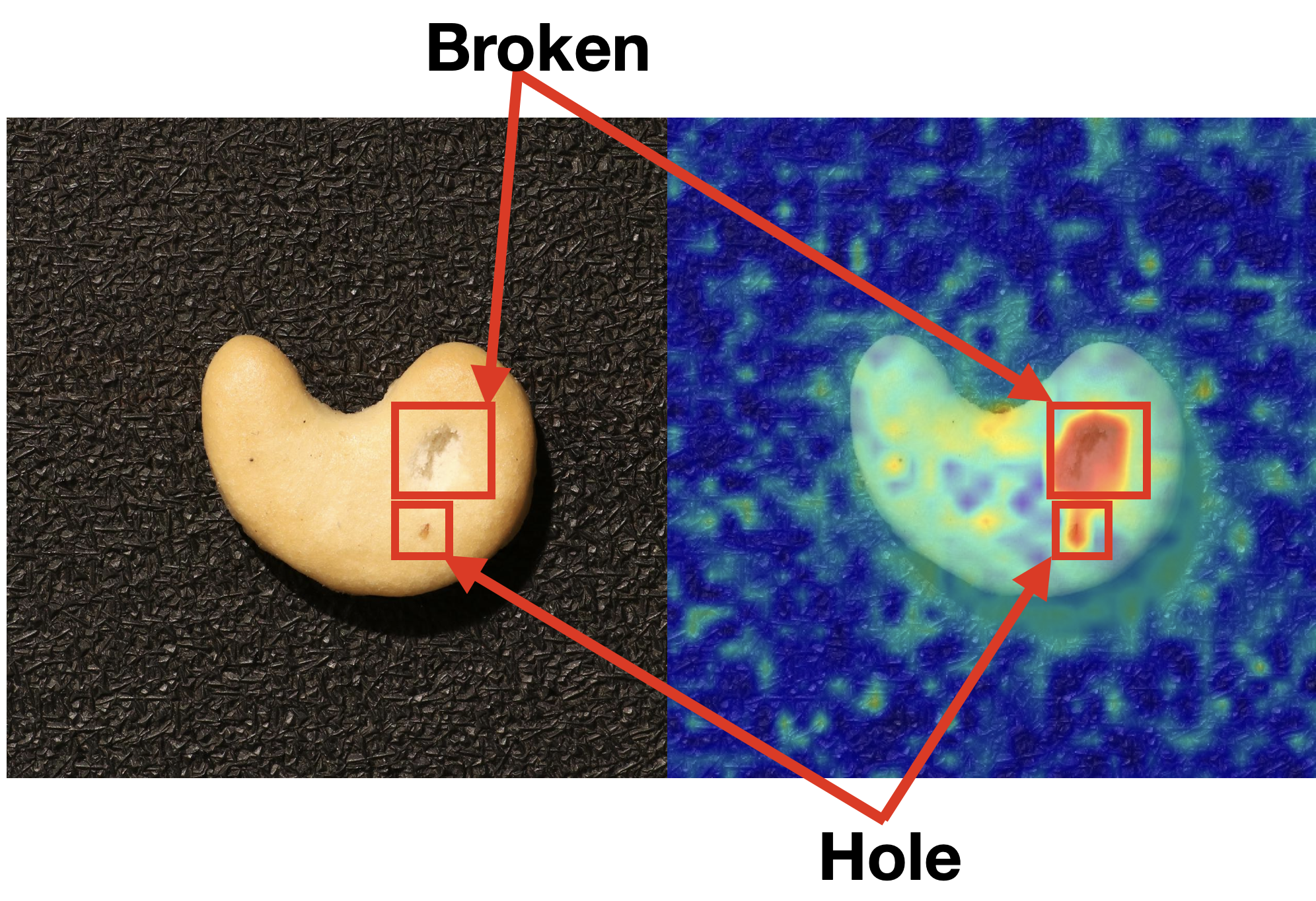}
        \caption{\small{Broken and Hole defects.}}
        \label{fig:subfig1}
    \end{subfigure}
    \hfill
    \begin{subfigure}[b]{0.22\textwidth}
        \centering
        \includegraphics[width=\textwidth]{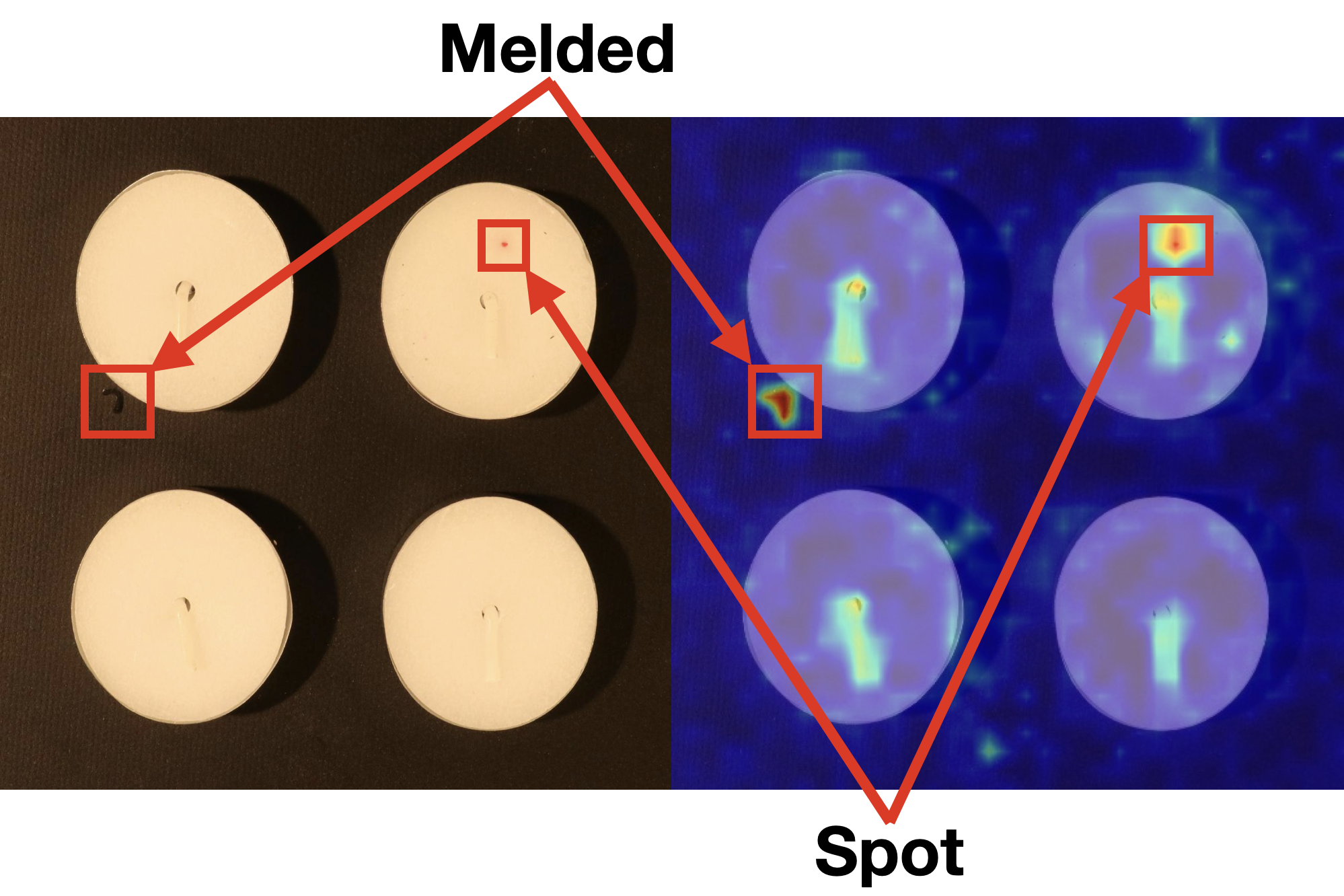}
        \caption{\small{Melded and Spot defects.}}
        \label{fig:subfig2}
    \end{subfigure}
    \caption{\small{MultiADS locates and identifies simultaneously multi-type anomalies on cashew (a) and candle (b) products.}}
    \label{fig:multi-seg_images}
    \vspace{-0.5em}
\end{figure}

\textbf{Multi-type Anomaly Awareness.} 
Figure~\ref{fig:multi-seg_images} shows that multiple defect types, such as \emph{broken} and \emph{hole}, can appear on one image, and MultiADS can successfully locate and classify these defects. 
Additionally, in Table~\ref{tab:multi-seg_tabels}, we listed the segmentation performance for some sample defect types that are seen/unseen during the training phase. 
We notice that defects such as \emph{holes} and \emph{damages} are relatively easy to locate and classify because they also occur on the training dataset - MVTec-AD. It may be that these defects are similar in terms of shape to those they have in datasets. 
For unseen defects like \emph{extra} and \emph{stuck}, our model achieves slightly lower accuracy. On the other hand, for other unseen defects such as \emph{pit}, we can still perform with high accuracy on the classification task. These results reflect that our approach has generalization ability even on large and complex datasets and unseen defects in the training dataset.

\begin{table}[t]
    \centering
    \caption{\small{Results on MTAS Task of MultiADS.}}\label{table:multi-class seg}
    \begin{adjustbox}{width=0.4\textwidth}
    \begin{tabular}{c|c|c c c}
        \toprule
        \multirow{2}{*}{\textbf{Train}}  & \multirow{2}{*}{\textbf{Target}}  & \multicolumn{3}{c}{Pixel-Level} \\
        \cline{3-5}
         & & AUROC & F1-score & AP \\
        \midrule
        \multirow{5}{*}{MVTec-AD}   & VisA  & 93.6 & 22.3 & 24.8 \\
         & MPDD & 95.2 & 42.8 & 53\\
         & MAD-sim & 92.1 & 27.9 & 31.5 \\
         & MAD-real & 89.2 & 52.5 & 52.3 \\
         & Real-IAD & 89.5 & 22.6 & 25.0 \\
        \midrule
        \multirow{2}{*}{VisA}  & MVTec-AD &  89.1 &24 & 30.5\\
        & MPDD &  95.3 & 46.7 & 50.5\\
        \midrule
        \multirow{2}{*}{MPDD}  & VisA &  93.4 & 22.1 & 23.3 \\
        & MVTec-AD &  89.4 & 23.9 & 27.6 \\
        \midrule
        \multirow{2}{*}{Real-IAD} & MVTec-AD &  87.7  & 21.4 &    29.9 \\
        & VisA &  88.1 & 23.8 &  24.8 \\
        \bottomrule
    \end{tabular}
    \end{adjustbox}
    \label{tab:multi-defect}
    \vspace{-1.45em}
\end{table}

\begin{table}[t]
    \centering
    \caption{\small{Results MTAS for zero-shot setting at pixel-level for sample defect-types. The model is trained on the MVTec-AD dataset. - indicates \textbf{unseen} defect types while \cmark indicates \textbf{seen} defect types during training.}}
    \begin{subtable}[t]{0.23\textwidth}
        \centering
        \caption{VisA}
        \begin{adjustbox}{width=0.99\textwidth}
        \begin{tabular}{p{0.3cm} c| c c c}
            \toprule
             & Defects & AUROC & F1-Score & AP  \\
            \hline
             - & Extra & 94.07 & 2.11 & 0.15  \\
             - & Stuck & 91.54 & 10.51 & 7.76 \\
             \cmark & Bent & 96.53 & 6.07 & 7.74  \\
             \cmark & Hole & 99.55 & 12.64 & 25.19 \\
            \bottomrule
        \end{tabular}
        \end{adjustbox}
    \end{subtable}
    \hfill
    \begin{subtable}[t]{0.23\textwidth}
        \centering
        \caption{Real-IAD}
        \begin{adjustbox}{width=0.99\textwidth}
        \begin{tabular}{p{0.3cm} c|c c c}
            \toprule
             & Defects & AUROC & F1-Score & AP  \\
            \hline
             - & Pit & 97.08 & 6.15 & 1.01  \\
             \cmark & Contamin. & 90.03 & 6.12 & 1.86   \\
             \cmark & Scratch & 92.63 & 4.37  & 2.96   \\
             \cmark & Damage &  96.61 & 6.31 & 9.75   \\
            \bottomrule
        \end{tabular}
        \end{adjustbox}
    \end{subtable}

    
    \label{tab:multi-seg_tabels}
    \vspace{-0.5em}
\end{table}

\textbf{Ablation Study.} We present the results of our ablation studies on MTAS, quantifying the contributions of the KBA component. As Table~\ref{tab:ablation_mtad} shows, the performance improves with the detailed text prompts constructed by KBA in both VisA and MAD-sim datasets. Similar patterns are present across all datasets. 

\begin{table}[ht]
    \centering
    \caption{\small{Ablation studies on the role of KBA for MTAS}}
    \resizebox{.42\textwidth}{!}{%
    \begin{tabular}{ c | c c c | c c c}
        \toprule
        & \multicolumn{3}{c|}{MVTec $\rightarrow$ VisA} & \multicolumn{3}{c}{MVTec $\rightarrow$ MAD-sim}\\
        \hline
        KBA & AUROC & F1-score & AP & AUROC & F1-score & AP\\
        \hline
        - & 87.0 & 22.1 & 23.6 & 91.1 & 25.1 & 26.5 \\
        \cmark & 93.6 & 22.3 & 24.8 & 92.1 & 27.9 & 31.5 \\
        \bottomrule
    \end{tabular}
    }
    \label{tab:ablation_mtad}
    \vspace{-1em}
\end{table}

\subsubsection{Binary Detection and Segmentation}

\textbf{ZSAD.}
In Table~\ref{table:resZeroVisA}, we show the performance on ZSAD for pixel-level (AUROC, AUPRO) and image-level (AUROC, AP) on VisA, MPDD, MAD (sim and real), and Real-IAD datasets. 
We selected these metrics to evaluate the performance following~\cite{AnomalyCLIP}. 
For a fair comparison, our approach and baseline approaches, including WinCLIP, April-GAN, AnomalyCLIP, and AdaCLIP, are trained on the MVTec-AD dataset.
We observe that MultiADS and MultiADS-F are the best overall performers, especially when performance is evaluated with the AUPRO and AUROC metrics at the pixel and image levels, respectively. We note that our approach achieves the best performance for all metrics on both levels on the recent datasets, MAD and Real-IAD, which are even more challenging. 
Meanwhile, MultiADS-F is the best overall performer on the MPDD, MAD-real, and Real-IAD datasets, indicating that text prompts of non-relevant defect types present more noise for these datasets. Note that MultiADS and MultiADS-F have the same scores for the MAD-sim dataset, as all defect types appear for all product types.
The best baseline performer is the AnomalyCLIP approach.  

\begin{table}[t]
\begin{center}
\caption{\small{Zero-shot anomaly detection and segmentation. (Bold represents best performer; underline indicates second best performer, * means results are taken from papers)}}\label{table:resZeroVisA}
\small
\scalebox{0.7}{
\begin{tabular}{ c|c|c|c c|c c } 
\toprule
\multicolumn{3}{c|}{\textbf{ZSAD}}  & \multicolumn{2}{c|}{Pixel-Level} & \multicolumn{2}{c}{Image-Level}\\ 

\hline
Dataset & Method  & Venue  & AUROC    & AUPRO & AUROC   & AP \\
 \hline		
\multirow{9}{*}{\textbf{VisA}} & CLIP* & \small{ICML21} & 46.6 &  14.8 & 66.4  & 71.5 \\
& CLIP-AC* & \small{ICML21} & 47.8 &  17.3 & 65.0  & 70.1  \\
& CoOp* & \small{IJCV22} & 24.2 & 3.8 & 62.8  & 68.1 \\
& CoCoOp* & \small{CVPR22} & 93.6 &  - & 78.1  & - \\
& WinCLIP & \small{CVPR23} & 79.6 &  56.8 & 78.1  & 81.2 \\
& April-GAN & \small{CVPR23} & 94.2 &  86.8 & 78.0   & 81.4 \\
& AnomalyCLIP & \small{CVPR24} & \textbf{95.5} &  {87.0} & {82.1} & {85.4} \\
& AdaCLIP & \small{ECCV24} & \underline{95} & - & 75.4 & 79.3 \\
\cline{2-7}	
& \multicolumn{2}{c|}{MultiADS (ours)}  & \underline{95}  & \textbf{89.7} & \textbf{83.6}  & \textbf{86.9} \\
& \multicolumn{2}{c|}{MultiADS-F (ours)}  & {94.5}  & \underline{87.4} & \underline{82.5}  & \underline{86.5} \\
\hline
\multirow{9}{*}{\textbf{MPDD}} & CLIP* & \small{ICML21} & 62.1 &  33.0 & 54.3 &  65.4 \\
& CLIP-AC* & \small{ICML21} & 58.7 &  29.1 & 56.2 & 66.0  \\
& CoOp* & \small{IJCV22} & 15.4 &  2.3 & 55.1 &  64.2 \\
& CoCoOp* & \small{CVPR22} & 95.2 &- & 61 & - \\
& WinCLIP & \small{CVPR23} & 76.4 &  48.9 & 63.6 & 69.9 \\
& April-GAN & \small{CVPR23} & 94.1 &  83.2 & 73.0 &  80.2 \\
& AnomalyCLIP & \small{CVPR24} & \textbf{96.5} &  88.7 & 77.0 &  \textbf{82.0} \\
& AdaCLIP & \small{ECCV24} & 96.3 & & 66.3 & 75 \\
\cline{2-7}	
& \multicolumn{2}{c|}{MultiADS (ours)}  & 95.8  & \textbf{89.7} & \underline{78.3} & 78.4 \\
& \multicolumn{2}{c|}{MultiADS-F (ours)}  & \underline{96.3}  & \underline{89.5} & \textbf{79.7} & \underline{80.5} \\
\hline
\multirow{6}{*}{\textbf{MAD-sim}} & WinCLIP & \small{CVPR23} & 77.6 & 55.8 & 54.3 & 90.2 \\
& April-GAN & \small{CVPR23} & 80.4 & 61.5 & 56 & 91 \\
& AnomalyCLIP & \small{CVPR24} & 77.9 & 40.1 & 54.6 & 90.9 \\
& AdaCLIP & \small{ECCV24} & 85.7 & - & 55.2 & 90.5 \\
\cline{2-7}	
& \multicolumn{2}{c|}{MultiADS (ours)}  & \multirow{2}{*}{\textbf{88.0}} & \multirow{2}{*}{\textbf{74.2}} & \multirow{2}{*}{\textbf{57.1}} & \multirow{2}{*}{\textbf{94.4}} \\
& \multicolumn{2}{c|}{MultiADS-F (ours)}  &  &  &  & \\

\hline
\multirow{6}{*}{\textbf{MAD-real}} & WinCLIP & \small{CVPR23} & 60.5 & 26.9 & 64.1 & 87.6 \\
& April-GAN & \small{CVPR23} & 88.2 & 69.5 & 62.9 & 87.7 \\
& AnomalyCLIP & \small{CVPR24} & {88.3} & 65.1 & 66.8 & 90 \\
& AdaCLIP & \small{ECCV24} & 85.7 & - & 59 & 86.5 \\
\cline{2-7}	
& \multicolumn{2}{c|}{MultiADS (ours)}  & \underline{89.7} & \underline{74.0} & \underline{78.3} & \textbf{92.9} \\
& \multicolumn{2}{c|}{MultiADS-F (ours)}  & \textbf{90.7} & \textbf{75.2} & \textbf{78.5} & \textbf{92.9} \\
\hline
\multirow{6}{*}{\textbf{Real-IAD}} & WinCLIP & \small{CVPR23} & 87.1 & 59.9 & 75 & 72.3 \\
& April-GAN & \small{CVPR23} & 96 &  86.8 & 75.7 & 73.5 \\
& AnomalyCLIP & \small{CVPR24} & 96.2 & 85.7 & \underline{78.4} & 76.7 \\
& AdaCLIP & \small{ECCV24} & 95.3 & - & 70.1 & 68.5 \\
\cline{2-7}	
& \multicolumn{2}{c|}{MultiADS (ours)}  & \textbf{96.6} & \underline{87.1} & \textbf{78.7} & \textbf{79.1} \\
& \multicolumn{2}{c|}{MultiADS-F (ours)}  & \underline{96.3} & \textbf{87.2} & 78.2 & \underline{78.5} \\
\bottomrule
\end{tabular}
}
\end{center}
\vspace{-1.5em}
\end{table}

Table \ref{tab:ablation_zsad} shows the ablation study quantifying the contributions of KBA, global anomaly score, and different stage numbers on the ZASD task. 
The stage number has the highest impact; the drop in performance is around $5\%$ in terms of AP for both datasets when $m=3$.


\begin{table}[ht]
    \centering
    \caption{\small{Ablation studies on the role of KBA, global anomaly score $a_x$, and stage number $m$ on the ZSAD task. Pixel-level results are ignored since $a_x$ is only used at the image-level.  }}
    \resizebox{.47\textwidth}{!}{%
    \begin{tabular}{c c c | c c c c | c c c c}
        \toprule
        \multicolumn{3}{c|}{\textbf{ZSAD}} & \multicolumn{4}{c|}{MVTec $\rightarrow$ VisA} & \multicolumn{4}{c}{MVTec $\rightarrow$ MPDD}\\
        \cline{4-11}
        & & & \multicolumn{2}{c|}{Pixel-Level} & \multicolumn{2}{c|}{Image-Level} & \multicolumn{2}{c|}{Pixel-Level} & \multicolumn{2}{c}{Image-Level}\\
        \hline
         $m$ & $a_x$ & KBA & AUROC & AUPRO & AUROC & AP  & AUROC & AUPRO & AUROC & AP\\
        \hline
        3 & \cmark & \cmark & 94.5 & 87.7 & 79.5 & 82.3 & 93.7 & 84.3 & 68.2 & 74.8  \\
        4 & - & \cmark & - & - & 82.1 & 85.8 & - & -  &  76.5 & 78.1\\
        4 & \cmark & - & 94.4 & 88.7 & 82.4 & 86.1 & 95.7 & 89.1 & 77.9 & 77.6\\
        \hline
        \rowcolor{gray!40} 4 & \cmark & \cmark & 95.0 & 89.7 & 83.6 & 86.9 & 95.8 & 89.5 & 78.3 & 78.4\\
        \bottomrule
    \end{tabular}
    }
    \label{tab:ablation_zsad}
\end{table}

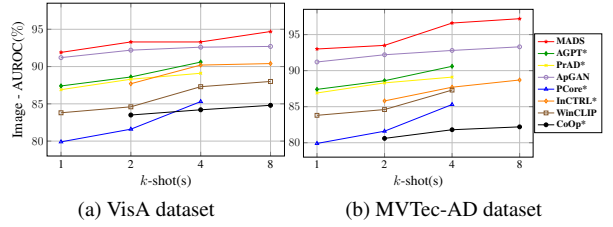
\begin{figure}[t]
\begin{center}
    \begin{subfigure}[b]{.2217\textwidth}
        \begin{adjustbox}{width=\textwidth}
            \begin{tikzpicture}
            \begin{axis}[
                width=10cm, height=7cm,
                xlabel={$k$-shot(s)},
                xlabel style={font=\Large},
                ylabel={Image - AUROC($\%$)},
                ylabel style={font=\Large},
                xmin=0.8, xmax=4.2,
                ymin=78, ymax=98.5,
                xtick={1,2,3,4},
                xticklabels={1,2,4,8},
                ticklabel style={font=\large},
                grid=major,
            ]

            \addplot[
                color=red,
                mark=star,
                thick
            ] coordinates { 
                (1,91.9)
                (2,93.3)
                (3,93.3)
                (4,94.7)
            };
        
            \addplot[
                color=green!60!black,
                mark=diamond*,
                thick
            ] coordinates { 
                (1,87.4)
                (2,88.6)
                (3,90.6)
            };
        
            \addplot[
                color=yellow,
                mark=x,
                thick
            ] coordinates { 
                (1,86.9)
                (2,88.3)
                (3,89.1)
            };
        
            \addplot[
                color=purple,
                mark=o,
                thick
            ] coordinates { 
                (1,91.2)
                (2,92.2)
                (3,92.6)
                (4,92.7)
            };
        
            \addplot[
                color=blue,
                mark=triangle,
                thick
            ] coordinates { 
                (1,79.9)
                (2,81.6)
                (3,85.3)
            };
        
            \addplot[
                color=orange,
                mark=diamond,
                thick
            ] coordinates { 
                (2,87.7)
                (3,90.2)
                (4,90.4)
            };
        
            \addplot[
                color=brown!70!black,
                mark=square,
                thick
            ] coordinates { 
                (1,83.8)
                (2,84.6)
                (3,87.3)
                (4,88)
            };
        
            \addplot[
                color=black,
                mark=*,
                thick
            ] coordinates { 
                (2,83.5)
                (3,84.2)
                (4,84.8)
            };
            \end{axis}
            \end{tikzpicture}
        \end{adjustbox}
        \caption{VisA dataset}\label{fig:fewshot_visa}
    \end{subfigure}%
    \hfill
    \begin{subfigure}[b]{.2535\textwidth}
    \begin{adjustbox}{width=\textwidth}
    \begin{tikzpicture}
    \begin{axis}[
        width=10cm, height=7cm,
        xlabel={$k$-shot(s)},
        xlabel style={font=\Large},
        xmin=0.8, xmax=4.2,
        ymin=78, ymax=98.5,
        xtick={1,2,3,4},
        xticklabels={1,2,4,8},
        ticklabel style={font=\large},
        legend style={
            font=\bfseries,
            legend columns=1,          
            at={(1.3,0.5)},            
            anchor=east,               
            /tikz/every even column/.append style={column sep=0.25em} 
        },
        legend cell align={left},
        grid=major,
    ]

    \addplot[
        color=red,
        mark=star,
        thick
    ] coordinates { 
        (1,93)
        (2,93.5)
        (3,96.6)
        (4,97.2)
    };
    \addlegendentry{MADS}

    \addplot[
        color=green!60!black,
        mark=diamond*,
        thick
    ] coordinates { 
        (1,87.4)
        (2,88.6)
        (3,90.6)
    };
    \addlegendentry{AGPT*}

    \addplot[
        color=yellow,
        mark=x,
        thick
    ] coordinates { 
        (1,86.9)
        (2,88.3)
        (3,89.1)
    };
    \addlegendentry{PrAD*}

    \addplot[
        color=purple,
        mark=o,
        thick
    ] coordinates { 
        (1,91.2)
        (2,92.2)
        (3,92.8)
        (4,93.3)
    };
    \addlegendentry{ApGAN}

    \addplot[
        color=blue,
        mark=triangle,
        thick
    ] coordinates { 
        (1,79.9)
        (2,81.6)
        (3,85.3)
    };
    \addlegendentry{PCore*}

    \addplot[
        color=orange,
        mark=diamond,
        thick
    ] coordinates { 
        (2,85.8)
        (3,87.7)
        (4,88.7)
    };
    \addlegendentry{InCTRL*}

    \addplot[
        color=brown!70!black,
        mark=square,
        thick
    ] coordinates { 
        (1,83.8)
        (2,84.6)
        (3,87.3)
    };
    \addlegendentry{WinCLIP}

    \addplot[
        color=black,
        mark=*,
        thick
    ] coordinates { 
        (2,80.6)
        (3,81.8)
        (4,82.2)
    };
    \addlegendentry{CoOp*}

    \end{axis}
\end{tikzpicture}
    \end{adjustbox}
    \caption{MVTec-AD dataset}\label{fig:few_shot_mvtec}
    \end{subfigure}
     \vspace{-1em}
    \caption{\small{Few-Shot Image level (AUROC) accuracy for different k-shots on the VisA and MVTec-AD datasets.  (* - results taken from papers, AGPT - AnomalyGPT, PCore - PatchCore, PrAD - PromptAD, ApGAN - April-GAN)}
    }
    \label{fig:few_shot_accuracy}
\end{center}
\vspace{-2em}
\end{figure}

\textbf{FSAD.}
Figure~\ref{fig:few_shot_accuracy} shows the results for the FSAD task, for image-level (AUROC) with different numbers of shots, $k = [1, 2, 4, 8]$, on the Visa and MVTec-AD datasets. Similarly to ZSAD, we train our model on the MVTec-AD dataset and test on VisA and vice versa. 
We note that the most competitive baselines are April-GAN, PromptAD, and AnomalyGPT. 
We observe that MultiADS is the best overall performer for both datasets. The same performance patterns are found on other datasets, too. The main advantage of our approach lies in extending the investigation based on defect awareness, supporting our claim that the main drawback of other methods is the two-state (normal and abnormal) limitation. 

Figure~\ref{fig:visualization} depicts a qualitative evaluation of the FSAD results of MultiADS and the best overall competitor, April-GAN, for scratch and hole defect types. 
We observe that MultiADS demonstrates higher confidence in identifying anomalies and achieves better segmentation across the same and different defect types due to its enhanced ability to capture the semantics of different defect types. More results are provided in the Appendix.



\input{tikz/visualization}

\section{Conclusion}
In this paper, we propose MultiADS, which constructs defect-aware text prompts to improve the performance of anomaly detection and segmentation tasks. We present a multi-type anomaly segmentation task that aims to determine the defect types and locations at the pixel level. 
We evaluated MultiADS on such a new task and positioned it as a baseline that can be used by the community. 
Finally, we evaluate MultiADS's performance against 12 baselines in ZSAD/FSAD on five datasets. 
Our evaluation demonstrates that MultiADS achieves the best performance in most cases for ZSAD/FSAD. 
In the future, we plan to explore adapting our approach to learn text prompt embeddings. 

\section{Acknowledgement}
This work was partially funded by the  European Union’s Horizon RIA research and innovation programme under grant agreement No. 101092908 (SMARTEDGE). The authors also thank the International Max Planck Research School for Intelligent Systems (IMPRS-IS) for supporting Hongkuan Zhou. 
{
    \small
    \bibliographystyle{ieeenat_fullname}
    \bibliography{main}
}
\clearpage

\maketitlesupplementary

\section{Our approach}

In this section, we will further discuss more details regarding our proposed approach, MultiADS. 

\subsection{Knowledge Base for Anomalies and Defect-Aware Text Prompts Design}
We construct text prompts based on the information we obtain from the Knowledge Base for Anomalies (KBA). 
This allows for leveraging the specificity of the defect type for each product class. 
The procedure for defect-aware prompt construction is consistently applied to each dataset. 
It should be noted, however, that the text prompt regarding the normal state and text template are the same for all datasets.

We conduct experiments on three commonly known datasets, namely MVTec-AD~\cite{MVTec}, VisA~\cite{VisA}, MPDD~\cite{MPDD}, MAD~\cite{PAD}, Real-IAD~\cite{realIAD}.
We construct multiple distinct defect-aware text prompts and $1$ for the normal state, for each dataset.
We construct text prompts that represent the normal or good state (without defects) of the images, using the following text prompt template:

\textit{normal = [
    ``\textit{[cls]}'', ``flawless [cls]'', ``perfect [cls]'', ``unblemished [cls]'', ``[cls] without flaw'', 
    ``[cls] without defect'', ``[cls] without damage'', 
    ``[cls] with immaculate quality'', ``[cls] without any imperfections'', '[cls] in ideal condition''
]}

where \textit{[cls]} represents a product class from a given dataset. 
We apply the same normal state design for all datasets, utilizing the text template as in~\cite{April-GAN} for all datasets as follows:

\textit{text-template = [``a bad photo of a \{\}.'',``a low resolution photo of the \{\}.'', ``a bad photo of the \{\}.'', ``a cropped photo of the \{\}.'', ``a bright photo of a \{\}.'', ``a dark photo of the \{\}.'', ``a photo of my \{\}.'', ``a photo of the cool \{\}.'', ``a close-up photo of a \{\}.'', ``a black and white photo of the \{\}.'', ``a bright photo of the \{\}.'', ``a cropped photo of a \{\}.'', ``a jpeg corrupted photo of a \{\}.'', ``a blurry photo of the \{\}.'', ``a photo of the \{\}.'', ``a good photo of the \{\}.'', ``a photo of one \{\}.'', ``a close-up photo of the \{\}.'', ``a photo of a \{\}.'', ``a low resolution photo of a \{\}.'', ``a photo of a large \{\}.'', ``a blurry photo of a \{\}.'', ``a jpeg corrupted photo of the \{\}.'', ``a good photo of a \{\}.'', ``a photo of the small \{\}.'', ``a photo of the large \{\}.'', ``a black and white photo of a \{\}.'', ``a dark photo of a \{\}.'', ``a photo of a cool \{\}.'', ``a photo of a small \{\}.'', ``this is a \{\} in the scene.'', ``this is the \{\} in the scene.'', ``this is one \{\} in the scene.'', ``there is the \{\} in the scene.'', ``there is a \{\} in the scene.'']}

where \{\} is filled with content from the normal and defect-aware text prompts.

An example of a text-prompt representing the normal state for product class \textit{[cls] = cable} is as follows:

\begin{multline}
S_{\text{normal}} =  \{``\text{A bad photo of \textit{cable}.}",  \\
             \cdots, \\
             ``\text{There is a \textit{cable in ideal condition} in the scene.}"\}
\end{multline}

Similarly, we construct text prompts representing distinct defect types. An example of a text-prompt representing the $bent$ defect type for product class \textit{[cls] = cable} is as follows:

\begin{multline}
S_{\text{bent}} =  \{ 
``\text{A bad photo of \textit{cable has a bent defect}.}",  \\
             \cdots, \\
             ``\text{There is a \textit{bent edge on cable} in the scene.}" 
             \} \\
\end{multline}

In Tables~\ref{tab:VISATextPrompts}-\ref{tab:RealIADTextPrompts}, we show the defect-aware text prompts for each defect type for all datasets, respectively. 
Note that for shared defect types among the datasets, such as \textit{bent}, \textit{hole}, and \textit{scratch}, we use the same defect-aware text prompts among all datasets. 

We provide the defined defect-aware text prompts, attached to the source code. 
The simplest way is to adapt the defect-aware information in a suitable manner based on the design of other approaches that aim to investigate defect types in anomaly detection tasks. 

In the main manuscript, we mention that the KBA contains the information for defect variations and defect type properties (attributes). Also, we include synonyms of defect types such as \textit{a slight curve},  which can also help VLMs to capture the similarity between image-text pairs. Likewise, we apply the same strategy for the construction of defect-aware text prompts for all defect types. More examples are provided in Tables~\ref{tab:VISATextPrompts}-\ref{tab:RealIADTextPrompts}.
Additionally, Tables~\ref{tab:ADMVTecADDetails}-\ref{tab:ADMRealIADDetails1} show variations of each defect type observed from all given datasets, for example \textit{bent} contains variations \textit{bent lead}, \textit{bent wire}, and \textit{bent edge}. 

\section{Datasets}
\begin{table}[ht]
    \caption{Key statistics on the datasets.}
    \begin{adjustbox}{width=0.4\textwidth}
    \begin{tabular}{c | c c c} 
        \toprule
        Dataset & Category & $|\mathcal{C}|$ & \makecell{Normal / Anomalous \\ Samples} \\ 
        \midrule
        MVTec-AD~\cite{MVTec} & \makecell{Object \\ Texture} & 15 & 4,096 / 1,258 \\
        \hline
        VisA~\cite{VisA} & Object & 12 & 9,621 / 1,200 \\
        \hline
        MPDD~\cite{MPDD} & Object & 6 & 1,064 / 282 \\
        MAD~\cite{PAD} & Object & 20 &  5,231 / 4,902\\
        Real-IAD~\cite{realIAD} & Object & 30 & 99,721 / 51,329 \\
        \bottomrule
    \end{tabular}
    \end{adjustbox}
    \label{tab:ADdatasets}
\end{table}
Due to space limitations in the main manuscript, here we describe in detail the industrial anomaly detection datasets:
MVTec-AD~\cite{MVTec}, VisA~\cite{VisA}, MPDD~\cite{MPDD}, MAD (simulated and real)~\cite{PAD}, and Real-IAD~\cite{realIAD}. 
Key statistics on the datasets are shown in Table~\ref{tab:ADdatasets}, such as categories, distinct classes, and the number of samples. 
MVTec-AD dataset consists of two categories, namely objects and textures, and $15$ product classes. 
For each product, there can be a different number of defects, as shown in Table~\ref{tab:ADMVTecADDetails}. 
This number varies from $1$ up to $8$, but for the textures, it is $5$ for all products. We classify each defect to the defect type as we defined before.

Additionally, we provide more details about defect types in order to highlight the importance and the design of our defect-aware text prompts. 
Thus, details of the VisA datasets are shown in Table~\ref{tab:ADVisADetails}; the products are categorized into complex structures, multiple instances (an image with multiple products of the same class, e.g., multiple candles, multiple capsules), and single instances. 
In total, it consists of $130$ defect types if we consider different combinations of defect types, but if we consider the combination as a single defect type, then the VisA dataset has $84$ defect types and $40$ distinct defect types. In Table~\ref{tab:ADVisADetails}, some defect types are included as part of the \textit{Combined} defect type, which consists of multiple defect types. The number of defect types for each product varies between $5$ and $9$ defect types.
In Table~\ref{tab:ADMPDDDetails}, we show detailed information regarding the MPDD dataset, which consists of $6$ product types and $11$ defect types, from which $8$ are distinct defect types. The number of defect types for each product varies between $1$ and $3$ defect types.
The MAD dataset consists of multi-pose views of twenty LEGO toys (product classes), with up to three anomaly types. It has simulated and real images. 
The Real-IAD dataset consists of thirty product categories, up to four defect types per category, and a larger proportion of defect area and range of defect ratios than other datasets. We utilize single-view image data. 
The details are illustrated in Table~\ref{tab:ADdatasets}. 

We apply the default normalization of CLIP~\cite{CLIP} to all datasets. After normalization, we resize the images to a resolution of $(518, 518)$ to obtain an appropriate visual feature map resolution. 

\begin{table*}[t]
    \centering
    \caption{Defect-Aware text prompts for all defect types of the VisA dataset. \textit{[cls]} represents a variable that takes as value all product classes in the VisA dataset. }
    \begin{adjustbox}{width=0.78\textwidth}
\begin{tabular}{c | c || c | c} 
        \toprule
        Defect Type & Defect-Aware Text Prompts & Defect Type & Defect-Aware Text Prompts\\ 
        \midrule
        Bent & \makecell{ ``\textit{[cls]} has a bent defect''  \\ ``flawed \textit{[cls]} with a bent lead'' \\ ``a bend found in \textit{[cls]}''  \\  ``\textit{[cls]} has a slight curve defect''  \\   ``\textit{[cls]} with noticeable bending'' \\ ``a bent wire on \textit{[cls]}'' } &
        Broken  & \makecell{ ``\textit{[cls]} with a breakage defect'' \\``broken \textit{[cls]}'' \\``\textit{[cls]} with broken defect'' \\``\textit{[cls]} shows breakage'' \\``broken or cracked areas on \textit{[cls]}'' \\``visible breakage on \textit{[cls]}``} \\
        \midrule
        Bubble  & \makecell{ ``\textit{[cls]} with bubbles defect'' \\``bubbles seen on \textit{[cls]}'' \\``\textit{[cls]} with bubble marks'' \\``air bubbles in \textit{[cls]}'' \\``\textit{[cls]} contains bubble defects'' \\``small bubbles on \textit{[cls]} surface``} &
        Burnt  & \makecell{ ``\textit{[cls]} with a burnt defect'' \\``\textit{[cls]} shows burn marks'' \\``burnt areas on \textit{[cls]}'' \\``\textit{[cls]} with signs of burning'' \\``scorch marks on \textit{[cls]}'' \\``\textit{[cls]} appears slightly burnt``} \\
        \midrule
        Chip  & \makecell{ ``\textit{[cls]} with chip defect'' \\``\textit{[cls]} with fragment broken defect'' \\``chipped areas on \textit{[cls]}'' \\``\textit{[cls]} with chipped parts'' \\``broken fragments on \textit{[cls]}'' \\``chip marks found on \textit{[cls]}``} & 
        Crack  & \makecell{ ``\textit{[cls]} with a crack defect'' \\``\textit{[cls]} has a visible crack'' \\``cracked areas on \textit{[cls]}'' \\``\textit{[cls]} with surface cracking'' \\``fine cracks found on \textit{[cls]}'' \\``\textit{[cls]} shows crack lines``} \\
\bottomrule
    \end{tabular}
    \end{adjustbox}
    \label{tab:VISATextPrompts}
    \vspace{4em}
\end{table*}


\begin{table*}[t]
    \centering
    \caption{Defect-Aware text prompts for all defect types of the MVTec-AD dataset. \textit{[cls]} represents a variable that takes as value all product classes in the MVTec-AD dataset. }
    \begin{adjustbox}{width=0.75\textwidth}
    \begin{tabular}{c | c || c | c} 
        \toprule
        Defect Type & Defect-Aware Text Prompts & Defect Type & Defect-Aware Text Prompts\\ 
        \midrule
     Bent & \makecell{ ``\textit{[cls]} has a bent defect''  \\ ``flawed \textit{[cls]} with a bent lead'' \\ ``a bend found in \textit{[cls]}''  \\  ``\textit{[cls]} has a slight curve defect''  \\   ``\textit{[cls]} with noticeable bending'' \\ ``a bent wire on \textit{[cls]}'' } &
    Broken & \makecell{``\textit{[cls]} has a broken defect''  \\ ``flawed \textit{[cls]} with breakage''  \\ ``visible breakage on \textit{[cls]}''  \\ ``\textit{[cls]} with broken areas''  \\ ``\textit{[cls]} shows signs of breaking''  \\ ``cracked or broken spots on \textit{[cls]}'' } \\
    \hline
    Color & \makecell{ ``\textit{[cls]} has a color defect''  \\ ``inconsistent color on \textit{[cls]}''  \\ ``\textit{[cls]} with color discrepancies'' \\  ``\textit{[cls]} has a noticeable color difference'' \\ ``\textit{[cls]} with irregular coloring'' \\  ``\textit{[cls]} has off-color patches'' } &
    Contamination & \makecell{ ``\textit{[cls]} has a contamination defect''  \\ ``foreign particles on \textit{[cls]}''  \\  ``\textit{[cls]} is contaminated'' \\ ``\textit{[cls]} contains contaminants'' \\ ``\textit{[cls]} has impurity issues'' \\  ``traces of contamination on \textit{[cls]}'' } \\
\bottomrule
    \end{tabular}
    \end{adjustbox}
    \label{tab:MVTecADTextPrompts}
\end{table*}

\begin{table*}[t]
    \centering
    \caption{Defect-Aware text prompts for all defect types of the MPDD dataset. \textit{[cls]} represents a variable that takes as value all product classes in the MPDD dataset. }
    \begin{adjustbox}{width=0.78\textwidth}
    \begin{tabular}{c | c || c | c} 
        \toprule
        Defect Type & Defect-Aware Text Prompts & Defect Type & Defect-Aware Text Prompts\\ 
        \midrule
        Bent & \makecell{ ``\textit{[cls]} has a bent defect''  \\ ``flawed \textit{[cls]} with a bent lead'' \\ ``a bend found in \textit{[cls]}''  \\  ``\textit{[cls]} has a slight curve defect''  \\   ``\textit{[cls]} with noticeable bending'' \\ ``a bent wire on \textit{[cls]}'' }
        &
        \makecell{Defective \\ Painting}  & \makecell{``\textit{[cls]} with a defective painting defect''\\ ``flawed \textit{[cls]} with painting imperfections'' \\ ``\textit{[cls]} has painting inconsistencies''\\ ``uneven painting on \textit{[cls]}'' \\ ``\textit{[cls]} shows poor paint quality''\\``paint defects present on \textit{[cls]}''}
        \\
        \hline
         Flattening  & \makecell{``\textit{[cls]} becomes flattened''\\``\textit{[cls]} has a flatten defect''\\ ``flattening observed on \textit{[cls]}'' \\ ``\textit{[cls]} appears compressed''\\``\textit{[cls]} is flattened or squashed''\\ ``deformation detected on \textit{[cls]}''} 
         &
        Hole  & \makecell{``\textit{[cls]} with a hole defect''\\ 'a hole on \textit{[cls]}''\\ 'visible hole in \textit{[cls]}'' \\ ``\textit{[cls]} with puncture marks''\\ 'hole detected in \textit{[cls]}''\\``\textit{[cls]} has small perforations''}
        \\
\bottomrule
    \end{tabular}
    \end{adjustbox}
    \label{tab:MPDDTextPrompts}
\end{table*}

\begin{table*}[t]
    \centering
    \caption{Defect-Aware text prompts for all defect types of the MAD dataset. \textit{[cls]} represents a variable that takes as value all product classes in the MAD dataset. }
    \begin{adjustbox}{width=0.78\textwidth}
    \begin{tabular}{c | c || c | c} 
        \toprule
        Defect Type & Defect-Aware Text Prompts & Defect Type & Defect-Aware Text Prompts\\ 
        \midrule
        Burr  & \makecell{``\textit{[cls]} has a burr defect''\\``sharp burr found on \textit{[cls]}''\\ ``\textit{[cls]} has excess material on edges'' \\ ``burr formation detected on \textit{[cls]}''\\``\textit{[cls]} exhibits rough edges''\\ `` \textit{[cls]} shows protruding material''} 
        &
        \makecell{Missing}  & \makecell{``\textit{[cls]} has a missing defect''\\ ``flawed \textit{[cls]} with something missing'' \\ ``\textit{[cls]} has missing components''\\ ``missing parts on \textit{[cls]}'' \\ ``\textit{[cls]} shows absent pieces''\\``certain parts missing from \textit{[cls]}''}
        \\
        \hline
        Stain & \makecell{ ``\textit{[cls]} with a stain defect''  \\ ``inconsistent color on \textit{[cls]}'' \\ ``\textit{[cls]} with color discrepancies''} 
        &  \multicolumn{2}{c}{} 
        \\
    \bottomrule
    \end{tabular}
    \end{adjustbox}
    \label{tab:MADTextPrompts}
\end{table*}

\begin{table*}[t]
    \centering
    \caption{Defect-Aware text prompts for all defect types of the Real-IAD dataset. \textit{[cls]} represents a variable that takes as value all product classes in the Real-IAD dataset. }
    \begin{adjustbox}{width=0.85\textwidth}
    \begin{tabular}{c | c || c | c} 
        \toprule
        Defect Type & Defect-Aware Text Prompts & Defect Type & Defect-Aware Text Prompts\\ 
        \midrule
        Pit & \makecell{ ``\textit{[cls]} has a pit defect''  \\ ``Small cavities or pits detected on \textit{[cls]}'' \\ ``\textit{[cls]} with color discrepancies''}
        &
        Scratch  & \makecell{``\textit{[cls]} has a scratch defect''\\ ``flawed \textit{[cls]} with a scratch'\\ 'scratches visible on \textit{[cls]}'' \\ ``\textit{[cls]} with surface scratches''\\ ``\textit{[cls]} has scratch marks''\\ ``minor scratches found on \textit{[cls]}''} 
        \\
        \hline
         Deformation  & \makecell{``\textit{[cls]} has a deformation defect''\\``\textit{[cls]} appears twisted or misshaped''\\ ``Structural distortion detected on \textit{[cls]}'' \\ ``Unexpected shape deformation found in \textit{[cls]}''\\``\textit{[cls]} exhibits rough edges''\\ `` \textit{[cls]} shows signs of bending under stress''} 
         &
         Deformation  & \makecell{``\textit{[cls]} has an abrasion defect''\\``\textit{[cls]} has noticeable or scuffing''\\ ``\textit{[cls]} is affected by continuous rubbing'' \\ ``Worn or scraped areas found on \textit{[cls]}''} 
         \\
         \hline
         Damaged & \makecell{ ``\textit{[cls]} has a damaged defect''  \\ ``flawed \textit{[cls]} with damage'' \\ ``\textit{[cls]} with visible damage'' \\ ``damaged areas on \textit{[cls]}''  \\ ``physical damage seen on \textit{[cls]}''  \\ ``noticeable wear on \textit{[cls]}'' }
         &
        Missing & \makecell{ ``\textit{[cls]} has a missing defect''  \\ ``flawed \textit{[cls]} with something missing'' \\ ``\textit{[cls]} has missing components'' \\ ``missing parts on \textit{[cls]}''  \\ ``\textit{[cls]} shows absent pieces'' \\ ``certain parts missing from \textit{[cls]}'' } 
        \\
\bottomrule
    \end{tabular}
    \end{adjustbox}
    \label{tab:RealIADTextPrompts}
\end{table*}

\begin{table*}[t]
    \begin{minipage}[t]{0.48\textwidth}
        \centering
        \caption{Detailed statistics on the MVTec-AD dataset.}
        \label{tab:ADMVTecADDetails}
        \begin{adjustbox}{width=\linewidth}
        \begin{tabular}{c| c| c| c| c | c} 
    \toprule
        \multirow{2}{*}{Category}  & \multirow{2}{*}{Product} & \multirow{2}{*}{Defects} & \multirow{2}{*}{Defect Type} & \multicolumn{2}{c}{Original Test} \\
        \cline{5-6}
         & & &  & Anomalous & Normal\\
        \midrule
        \multirow{48}{*}{Objects}  & Bottle & \makecell{Broken Large\\ Broken Small\\ Contamination} & \makecell{Broken \\ Broken\\ Contamination}  & \makecell{20\\22\\21} & 20  \\
        \cline{2-6}
         & Cable & \makecell{Bent Wire\\ Cable Swap\\ Combined\\ Cut Inner Insulation\\ Cut Outer Insulation\\ Missing Cable\\ Missing Wire\\ Poke Insulation} & \makecell{Bent \\ Misplaced\\ Combined\\ Cut \\ Cut \\ Missing \\ Missing \\ Poke }  &\makecell{13\\ 12\\ 11\\ 14\\ 10\\ 12\\ 10\\ 10} & 58   \\
        \cline{2-6}
         & Capsule & \makecell{Crack\\ Faulty Imprint\\ Poke\\ Scratch\\ Squeeze} & \makecell{Crack\\ Faulty Imprint\\ Poke\\ Scratch\\ Squeeze}  & \makecell{23\\ 22\\ 21\\ 23\\ 20} & 23    \\
        \cline{2-6}
         
         & Hazelnut & \makecell{Crack\\ Cut\\ Hole\\ Print} & \makecell{Crack\\ Cut\\ Hole\\ Faulty Imprint} &\makecell{18\\ 17\\ 18\\ 17}  & 40   \\
        \cline{2-6}
         & Metal Nut & \makecell{Bent\\ Color\\ Flip\\ Scratch} & \makecell{Bent\\ Color\\ Misplaced\\ Scratch} & \makecell{25\\ 22\\ 23\\ 23} & 22   \\
        \cline{2-6}
         & Pill & \makecell{Color\\ Combined\\ Contamination\\ Crack\\ Faulty Imprint\\ Pill Type\\ Scratch} & \makecell{Color\\ Combined\\ Contamination\\ Crack\\ Faulty Imprint\\ Damaged\\ Scratch} &\makecell{25\\ 17\\ 21\\ 26\\19\\9\\24}  &26   \\
\bottomrule
    \end{tabular}
        \end{adjustbox}
    \end{minipage}
    \hfill 
    \hfill
    \hfill
    \hfill
    \begin{minipage}[t]{0.49\textwidth}
        \centering
        \caption{Detailed statistics on the VisA dataset. We relabeled every image originally marked as “combined” in the VisA dataset by identifying each individual defect it contains and assigning the image to all corresponding defect categories.}
        \label{tab:ADVisADetails}
        \begin{adjustbox}{width=\linewidth}
        \begin{tabular}{c| c| c| c| c | c} 
        \toprule
        \multirow{2}{*}{Category}  & \multirow{2}{*}{Product} & \multirow{2}{*}{Defects} &\multirow{2}{*}{Defect Type} & \multicolumn{2}{c}{Test} \\
        \cline{5-6}
         & &  &  & Anomalous & Normal \\
        \midrule
        \multirow{19}{*}{\makecell{Complex\\Structure}}
                               & \multirow{4}{*}{Pcb1}  & Bent       & Bent     &    15 & \multirow{4}{*}{100} \\
                               &       & Melt       & Melt     &    52 &                      \\
                               &       & Missing    & Missing  &    20 &                      \\
                               &       & Scratch    & Scratch  &    21 &                      \\
        \cline{2-6}
        & \multirow{4}{*}{Pcb2}
                 & Bent         & Bent         &    15 & \multirow{4}{*}{100} \\
        &         & Melt         & Melt         &    54 &                      \\
\bottomrule
    \end{tabular}
        \end{adjustbox}
    \end{minipage}
\end{table*}

\begin{table}[t]
    \centering
    \caption{Detailed statistics on the MPDD dataset.}
    \begin{adjustbox}{width=0.49\textwidth}
    \begin{tabular}{c| c| c| c | c} 
        \toprule
         \multirow{2}{*}{Product} & \multirow{2}{*}{Defects} & \multirow{2}{*}{Defect Type} & \multicolumn{2}{c}{Original Test} \\
        \cline{4-5}
          &  &   & Anomalous & Normal\\
        \midrule
         Bracket Black & \makecell{Hole\\Scratches} & \makecell{Hole\\Scratch}  & \makecell{12\\35}    & 32 \\
        \cline{1-5}         
         Bracket Brown& \makecell{Bend Mismatch\\ Parts Mismatch} & \makecell{ Mismatch\\ Mismatch}&\makecell{17\\45} & 26    \\
        \cline{1-5}
         Bracket White & \makecell{Defective Painting\\Scratches} & \makecell{Defective Painting\\Scratch}& \makecell{13\\17}& 30     \\
        \cline{1-5}
         Connector & Parts Mismatch & Mismatch  & 14 & 30    \\
        \cline{1-5}
         Metal Plate & \makecell{Major Rust\\Scratches\\Total Rust} & \makecell {Rust\\Scratch \\ Rust}  & \makecell{14\\34\\23}  & 26   \\
        \cline{1-5}
         Tubes & Anomalous & Flattening  & 69  & 32   \\
        \bottomrule
    \end{tabular}
    \end{adjustbox}
    \label{tab:ADMPDDDetails}
\end{table}

\begin{table}[t]
    \centering
    \caption{Detailed statistics on the MAD-real dataset.}
    \begin{adjustbox}{width=0.4\textwidth}
    \begin{tabular}{c| c| c| c  c} 
        \toprule
         \multirow{2}{*}{Product} & \multirow{2}{*}{Defects} & \multirow{2}{*}{Defect Type} & \multicolumn{2}{c}{Original Test} \\
        \cline{4-5}
          &  &  & Anomalous & Normal \\
        \midrule
         Bear & Stains & Stains  & 24 & 5    \\
        \cline{1-5}         
         Bird & Missing & Missing  & 22 & 5    \\
        \cline{1-5}
        Elephant & Missing & Missing  & 18 & 5    \\
        \cline{1-5}
        Parrot & Missing & Missing  & 23  & 5   \\
        \cline{1-5}
         Puppy & Stains & Stains  & 20 & 5    \\
        \cline{1-5}
         Scorpion & Missing & Missing  & 23 & 5    \\
        \cline{1-5}
         Turtle & Stains & Stains  & 21 & 5    \\
        \cline{1-5}
         Unicorn & Missing & Missing  & 21 & 5    \\
        \cline{1-5}
         Whale & Stains & Stains  & 32  & 5   \\
        \bottomrule
    \end{tabular}
    \end{adjustbox}
    \label{tab:ADMADrealDetails}  
\end{table}

\begin{table}[t]
    \centering
    \caption{Detailed statistics on the MAD-sim dataset.}
    \scalebox{0.7}{    
    \begin{tabular}{c| c| c| c | c} 
        \toprule
        \multirow{2}{*}{Product} & \multirow{2}{*}{Defects} & \multirow{2}{*}{Defect Type} & \multicolumn{2}{c}{Original Test} \\
        \cline{4-5}
          &  &  & Anomalous  & Normal\\
        \midrule
        Bear & \makecell{Burrs \\ Missing \\ Stains} & \makecell{Burrs \\ Missing \\ Stains}  & \makecell{88 \\ 112 \\ 59 } & 36\\ 
         \cline{1-5}
        Bird & \makecell{Burrs \\ Missing \\ Stains} & \makecell{Burrs \\ Missing \\ Stains}  & \makecell{51 \\ 160 \\ 40 } & 30\\ 
         \cline{1-5}
        Cat & \makecell{Burrs \\ Missing \\ Stains} & \makecell{Burrs \\ Missing \\ Stains}  & \makecell{98 \\ 151 \\ 58 } & 36\\ 
         \cline{1-5}
        Elephant & \makecell{Burrs \\ Missing \\ Stains} & \makecell{Burrs \\ Missing \\ Stains}  & \makecell{72 \\ 149 \\ 55 } & 36 \\ 
         \cline{1-5}
        Gorilla & \makecell{Burrs \\ Missing \\ Stains} & \makecell{Burrs \\ Missing \\ Stains}  & \makecell{67 \\ 137 \\ 35 } & 20\\ 
         \cline{1-5}
        Mallard & \makecell{Burrs \\ Missing \\ Stains} & \makecell{Burrs \\ Missing \\ Stains}  & \makecell{27 \\ 157 \\ 33 }& 20 \\
\bottomrule
    \end{tabular}
    }
    \label{tab:ADMADsimDetails}
\end{table}

\begin{table*}[t]
    \begin{minipage}[t]{0.48\textwidth}
        \centering
        \caption{Detailed statistics on the Real-IAD dataset (Part I).}
        \label{tab:ADMRealIADDetails1}
        \begin{adjustbox}{width=\linewidth}
        \begin{tabular}{c| c| c| c c} 
        \toprule
        \multirow{2}{*}{Product} & \multirow{2}{*}{Defects} & \multirow{2}{*}{Defect Type} & \multicolumn{2}{c}{Original Test} \\
        \cline{4-5}
          &  &  & Normal & Anomalous \\
        \midrule
         Audiojack & \makecell{Deformation \\ Scratch \\ Missing \\ Contamination } & \makecell{Deformation \\ Scratch \\ Missing \\ Contamination } & 398 & \makecell{126 \\ 4 \\ 56 \\ 27} \\ 
        \cline{1-5}
        Bottle Cap & \makecell{Pit \\ Scratch \\ Missing Parts \\ Contamination } & \makecell{Pit \\ Scratch \\ Missing Parts \\ Contamination } & 369 & \makecell{65 \\ 125 \\ 1 \\ 73} \\ 
        \cline{1-5}
        Button Battery & \makecell{Pit \\ Abrasion \\ Scratch \\ Contamination } & \makecell{Pit \\ Abrasion \\ Scratch \\ Contamination } & 291 & \makecell{123 \\ 68 \\ 109 \\ 117} \\ 
        \cline{1-5}
        End Cap & \makecell{Scratch \\ Damage \\ Missing Parts \\ Contamination } & \makecell{Scratch \\ Damage \\ Missing Parts \\ Contamination } & 289 & \makecell{92 \\ 119 \\ 133 \\ 80} \\ 
        \cline{1-5}
        Eraser & \makecell{Pit \\ Scratch \\ Missing Parts \\ Contamination } & \makecell{Pit \\ Scratch \\ Missing Parts \\ Contamination } & 389 & \makecell{36 \\ 101 \\ 30 \\ 68} \\ 
        \cline{1-5}
        Fire Hood & \makecell{Pit \\ Scratch \\ Missing Parts \\ Contamination } & \makecell{Pit \\ Scratch \\ Missing Parts \\ Contamination } & 418 & \makecell{33 \\ 51 \\ 62 \\ 23} \\
\bottomrule
    \end{tabular}
        \end{adjustbox}
    \end{minipage}
    \hfill 
    \hfill
    \hfill
    \hfill
    \begin{minipage}[t]{0.49\textwidth}
        \centering
        \caption{Detailed statistics on the Real-IAD dataset (Part II).}
        \label{tab:ADMRealIADDetails2}
        \begin{adjustbox}{width=\linewidth}
        \begin{tabular}{c| c| c| c c} 
        \toprule
        \multirow{2}{*}{Product} & \multirow{2}{*}{Defects} & \multirow{2}{*}{Defect Type} & \multicolumn{2}{c}{Original Test} \\
        \cline{4-5}
          &  &  & Normal & Anomalous \\
        \midrule
        Terminalblock & \makecell{Pit \\ Missing Parts \\ Contamination } & \makecell{Pit \\ Missing Parts \\ Contamination } & 308 & \makecell{142 \\ 145 \\ 106} \\ 
        \cline{1-5}
        Toothbrush & \makecell{Abrasion \\ Missing Parts \\ Contamination } & \makecell{Abrasion \\ Missing Parts \\ Contamination } & 272 & \makecell{170 \\ 137 \\ 149} \\ 
        \cline{1-5}
        Toy & \makecell{Pit \\ Scratch \\ Missing Parts \\ Contamination } & \makecell{Pit \\ Scratch \\ Missing Parts \\ Contamination } & 250 & \makecell{125 \\ 127 \\ 126 \\ 126} \\ 
        \cline{1-5}
        Toy-brick & \makecell{Pit \\ Scratch \\ Missing Parts \\ Contamination } & \makecell{Pit \\ Scratch \\ Missing Parts \\ Contamination } & 370 & \makecell{67 \\ 60 \\ 81 \\ 53} \\ 
        \cline{1-5}
        Transistor1 & \makecell{Deformation \\ Missing Parts \\ Contamination } & \makecell{Deformation \\ Missing Parts \\ Contamination } & 265 & \makecell{171 \\ 164 \\ 134} \\ 
        \cline{1-5}
        U Block & \makecell{Abrasion \\ Scratch \\ Missing Parts \\ Contamination } & \makecell{Abrasion \\ Scratch \\ Missing Parts \\ Contamination } & 436 & \makecell{20 \\ 17 \\ 44 \\ 45} \\
\bottomrule

    \end{tabular}
        \end{adjustbox}
    \end{minipage}
\end{table*}

\clearpage

\section{Baselines}

To demonstrate the performance of MultiADS, we compare MultiADS with broad SOTA baselines. We run experiments for April-GAN~\cite{April-GAN}, and other baseline results are taken from original papers. If the baseline does not report results for a specific dataset, then the results are taken from the latest publication, which includes these results. 
Details regarding each baseline are given as follows:

\begin{itemize}
    \item PaDiM~\cite{PaDiM} utilizes a pre-trained Convolutional Neural Network (CNN) for patch embedding and multivariate Gaussian distributions to get a probabilistic representation for a one-class learning setting, the normal class. Also, it considers the semantic relations of CNN to improve the localization. Results are taken from \cite{April-GAN, GraphCore} baselines. Source code is available at \url{https://github.com/taikiinoue45/PaDiM}.
    \item CLIP~\cite{CLIP} is a powerful zero-shot classification method. Results are taken from \cite{AnomalyCLIP} baseline, and to perform the anomaly detection task, they use two classes of text prompt templates "\textit{A photo of a normal [cls]}" and "\textit{A photo of an anomalous [cls]}", where "\textit{cls}" denotes the target class name. The anomaly score is computed according to Eq. [1] in the main manuscript. As for anomaly segmentation, they extend the above computation to local visual embedding to derive the segmentation. Source code is available at \url{https://github.com/openai/CLIP}.
    \item CLIP-AC~\cite{CLIP} employs an ensemble of text prompt templates that are recommended for the ImageNet dataset~\cite{CLIP}. Results are taken from \cite{AnomalyCLIP} baseline, and they average the generated textual embeddings of normal and anomaly classes, respectively, and compute the probability and segmentation in the same way as CLIP. Source code is available at \url{https://github.com/openai/CLIP}.
    \item RegAD~\cite{RegAD} is a few-shot learning approach that leverages feature registration as a category-agnostic approach. This approach trains a single generalizable model and does not require re-training or parameter fine-tuning for new categories. Results are taken from the original publication. Source code is available at \url{https://github.com/MediaBrain-SJTU/RegAD}.
    \item CoOp~\cite{CoOp} is a representative method for prompt learning. Results are taken from \cite{AnomalyCLIP} baseline for zero-shot setting and from \cite{InCTRL} for few-shot setting. To adapt CoOp to zero- and few-shot anomaly detection, authors of \cite{AnomalyCLIP, InCTRL} replace its learnable text prompt templates $[V_1][V_2]\dots[V_N][cls]$ with normality and abnormality text prompt templates, where $V_i$ is the learnable word embeddings. The normality text prompt template is defined as $[V_1][V_2]...[V_N][normal][cls]$, and the abnormality one is defined as $[V_1][V_2]\dots[V_N][anomalous][cls]$. Anomaly probabilities and segmentation are obtained in the same way as for AnomalyCLIP, and all parameters are kept the same as in the original paper. Source code is available at \url{https://github.com/KaiyangZhou/CoOp}.
    \item CoCoOp~\cite{CoCoOp} extends the CoOp work by generalizing the learned context to wider unseen classes within the same dataset. CoCoOp learns a lightweight neural network to generate for each image an input-conditional token (vector), and the proposed dynamic prompts adapt to each instance and are less sensitive to class shift. Results are taken from \cite{AnomalyCLIP} baseline.
    Source code is available at \url{https://github.com/KaiyangZhou/CoOp}.
    \item PatchCore~\cite{PatchCore} utilizes locally aggregated, mid-level patch features over a local neighborhood to ensure the retention of sufficient spatial context. PatchCore employs a memory bank for patch features to leverage nominal context at test time by using a greedy coreset subsampling. Results are taken from \cite{April-GAN} baseline. Source code is available at \url{https://github.com/amazon-science/patchcore-inspection}
    \item WinCLIP~\cite{WinCLIP} is a SOTA zero-shot anomaly detection method. Results for zero-shot settings are taken from the original publication and for few-shot settings are taken from ~\cite{April-GAN} baseline. The authors design a large set of text prompt templates specific to anomaly detection and use a window scaling strategy to obtain anomaly segmentation. Source code is available at \url{https://github.com/caoyunkang/WinClip}.
    \item April-GAN~\cite{April-GAN} is an improved version of WinCLIP. We conducted experiments with this approach and all parameters are kept the same as in their paper. April-GAN first adjusts the text prompt templates and then introduces learnable linear projections to improve local visual semantics to derive more accurate segmentation. Source code is available at \url{https://github.com/ByChelsea/VAND-APRIL-GAN}.
    \item GraphCore~\cite{GraphCore} is a few-shot learning approach that utilizes memory banks to store image features. Results are taken from the original publication. They employ graph representation (Graph Neural Networks) to provide a visual isometric invariant feature (VIIF) as an anomaly measurement feature. The VIIF reduces the size of redundant features stored in memory banks. Results are taken from the original publication. The authors have not provided a link to the source code yet.
    \item FastRecon~\cite{FastRecon} is a few-shot learning approach that utilizes a few normal samples as a reference to reconstruct its normal version, and sample alignment helps to detect anomalies. Thus, they propose a regression algorithm with distribution regularization for the transformation estimation. Results are taken from the original publication. Source code is available at \url{https://github.com/FzJun26th/FastRecon}.
    \item InCTRL~\cite{InCTRL} is a vision-language few-shot learning model that proposes an in-context residual learning approach. It aims to distinguish anomalies from normal samples by detecting residuals between test images and in-context few-shot normal sample prompts from the target domain on the fly. Results are taken from the original publication. Source code is available at \url{https://github.com/mala-lab/InCTRL}.
    \item PromptAD~\cite{PromptAD} is a vision-language few-shot learning approach that learns text prompts for anomaly detection. They propose to concatenate anomaly suffixes to transpose the semantics of normal prompts, in order to construct negative samples. They aim to control the distance between normal and abnormal prompt features through a hyperparameter. Results are taken from the original publication. Source code is available at \url{https://github.com/FuNz-0/PromptAD}.
    \item AnomalyCLIP~\cite{AnomalyCLIP} is a SOTA zero-shot anomaly detection method. Results are taken from the original publication. This approach learns a vector representation for text prompts for two states: normal and abnormal. They construct two templates of text prompts, object-aware text prompts and object-agnostic text prompts templates. 
    Through an object-agnostic text prompt template, they aim to learn the shared patterns of different anomalies. 
    Results are taken from the original publication. Source code is available at \url{https://github.com/zqhang/AnomalyCLIP}.
\end{itemize}

\section{Experiments}

In this section, we provide more details regarding our approach through ablation studies and the experiments that were conducted. We also visualize the results and discuss some insights and limitations of our approach.

\begin{table*}[t]
\begin{center}
\caption{Ablation study for testing without global anomaly score. MultiADS is our proposed method, while MultiADS-L is the ablated version without including the global anomaly score.}
\label{table:ablation_study_global_anomaly_score}
\small
\begin{tabular}{ c|c|c|c c c } 
\hline
Settings & \multirow{2}{*}{Training $\rightarrow$ Testing} & \multirow{2}{*}{Method} & \multicolumn{3}{c}{Image-Level}\\ 
\cline{4-6}
& & & AUROC & F1-max  & AP \\	
\hline
\multirow{4}{*}{Zero-shot} & \multirow{2}{*}{MVTec-AD $\rightarrow$ VisA} & MultiADS  & 83.6 & 80.3 & 86.9 \\
& & MultiADS-L & 82.1 \textcolor{light-green}{(+1.5)} & 80.3 \textcolor{light-green}{(+0.0)} & 85.8 \textcolor{light-green}{(+1.1)} \\
\cline{2-6}
& \multirow{2}{*}{MVTec-AD $\rightarrow$ MPDD} & MultiADS & 78.3 & 79.2 & 78.4 \\
& & MultiADS-L & 76.5 \textcolor{light-green}{(+1.8)} & 79 \textcolor{light-green}{(+0.2)} & 78.1 \textcolor{light-green}{(+0.3)} \\
\hline
\multirow{4}{*}{Few-shot (k=4)} & \multirow{2}{*}{MVTec-AD $\rightarrow$ VisA} & MultiADS & 93.3 & 89.7 & 94.3 \\
& & MultiADS-L & 93.8 \textcolor{light-red}{(-0.5)} & 89.6 \textcolor{light-green}{(+0.1)} & 94.5 \textcolor{light-red}{(-0.2)} \\
\cline{2-6}
& \multirow{2}{*}{MVTec-AD $\rightarrow$ MPDD} & MultiADS & 86 & 87.2 & 89.4 \\
& & MultiADS-L & 85.6 \textcolor{light-green}{(+0.4)}& 86.8 \textcolor{light-green}{(+0.4)}& 89.3 \textcolor{light-green}{(+0.1)} \\
\hline
\end{tabular}
\end{center}
\end{table*}

\begin{table*}[t]
\begin{center}
\caption{Ablation Study: Results for MultiADS for each product of the MPDD dataset with different defect-aware text prompts from the VisA dataset and the MPDD dataset on few-shot (k=1) anomaly detection and segmentation tasks. Our model is trained on the MVTec-AD dataset. (\textbf{Bold} represents the best performer)}\label{tab:AblationStudyMPDDVisA}
\small
\begin{adjustbox}{width=1\textwidth}
\begin{tabular}{ c| c c | c c|c c | c c| c c| c c| c c} 
\hline
Setting & \multicolumn{14}{c}{k=1}  \\
\hline
\textbf{MVTec $\rightarrow$ MPDD}  & \multicolumn{8}{c|}{Pixel-Level} & \multicolumn{6}{c}{Image-Level} \\ 
\hline
 \multirow{2}{*}{Product}    & \multicolumn{2}{c|}{AUROC}  & \multicolumn{2}{c|}{F1-max} & \multicolumn{2}{c|}{AP} & \multicolumn{2}{c|}{AUPRO} & \multicolumn{2}{c|}{AUROC} & \multicolumn{2}{c|}{F1-max}  & \multicolumn{2}{c}{AP}\\
 & VisA & MPDD & VisA & MPDD & VisA & MPDD & VisA & MPDD & VisA & MPDD & VisA & MPDD & VisA & MPDD \\
 \hline
 Bracket\_black 	&	96.7	&	97.2	&	11.2	&	18.7	&	4.5	&	11.8	&	88	&	89.5	&	63.4	& 74.6	&	78.5	&	81.6	&	68.6	&	80.8	\\
Bracket\_brown 	&	96	&	96.2	&	14.9	&	17.6	&	7.5	&	8.7	&	91	&	91.1	&	60.4	&	53.3	&	80&	79.7	&	72.5	&	71.4	\\
Bracket\_white 	&	99.7	&	99.7	&	20.7	&	24.5&	12.8	&	15.2	&	96.5	&	96.7	&	73.4	&	81.1	&	75	&	78.3	&	77	&	82.5	\\
Connector    & 		95.9	&	96.4	&	35.3	&	33.9	&	33.7	&	32.4	&	87.2	&	87.8	&	92.9	&	91.4	&	78.8	&	82.8	&	88.9	&	9.3	\\
Metal\_plate   &		96.3	&	96.3	&	74.6	&	73.1	&	81.2	&	74.8	&	90.6	&	89.8	&	99	&	92	&	97.9	&	90.1	&	99.6	&	97.2	\\
Tubes    &     		98.7	&	98.8	&	69	&	68.7	&	71	&	70.4	&	95	&	95.5	&	97.3	&	97.6	&	96.4	&	95.5	&	99	&	99.1	\\
\hline																													
Average    &      		97.2	&	\textbf{97.4}	&	37.6	&	\textbf{39.4}	&	35.1	&	\textbf{35.6}	&	91.4	&	\textbf{91.7}	&	81.1	&	\textbf{81.7}	&	84.4	&	\textbf{84.6}	&	84.3	&	\textbf{86.7}	\\
\hline
\end{tabular}
\end{adjustbox}
\end{center}
\end{table*}

\subsection{Experiment Details}
In this subsection, we detail the experimental setup. We use the ViT-L-14-336 CLIP backbone from OpenCLIP~\cite{ilharco_gabriel_2021_5143773}, pre-trained on the LAION-400M\_E32 setting of open-clip. The learning rate is set to 0.001, with a batch size of 8. The stage number $m=4$. The features are selected from layers 6, 12, 18, and 24.

We adopt a transfer learning setting, training the model on one dataset and evaluating it on the remaining. Specifically, we train our model on MVTec-AD and evaluate it on VisA, MPDD, MAD, and Real-IAD, as well as train on VisA and evaluate on MVTec-AD. Other combinations are not included in the results, as most baselines focus on the aforementioned configurations. During training, we exclude all images labeled with “combined” defects, which indicate multiple defects in a single image. This exclusion is due to the datasets providing binary anomaly masks that treat all defects as identical. Since combined defects are relatively rare in the datasets (see Tables~\ref{tab:ADMVTecADDetails}, \ref{tab:ADVisADetails}, \ref{tab:ADMPDDDetails}), we opted to leave them out during training. However, for testing, all images with multiple defects are included to ensure a fair comparison.

\subsection{Ablation Studies}

Here, we will give more details regarding our ablation studies and show additional results of the experiments we have conducted for the multi-type anomaly segmentation (MTAS) task, binary zero-/few-shot anomaly detection task, and zero-batch task. 

\subsubsection{Global Anomaly Score}

To assess the impact of the global anomaly score on anomaly detection, we conducted ablation studies using our MultiADS model without the global anomaly score, referred to as MultiADS-L. As shown in Table~\ref{table:ablation_study_global_anomaly_score}, removing the global anomaly score leads to a noticeable performance drop in the zero-shot setting. However, the performance drop in the few-shot setting is minimal, likely because the additional information provided by the test data compensates for the absence of global context.

\subsubsection{Defect-Aware Text Prompts}

To show the importance of the defect-aware text prompts, we conduct experiments on the MPDD dataset with our approach, MultiADS. First, we train our model on the MVTec-AD dataset, with defect-aware text prompts constructed for the MVTec-AD dataset. Then, during the testing phase, instead of using the defect-aware text prompts constructed for the MPDD dataset, we use defect-aware text prompts constructed for the VisA dataset. The results are shown in Table~\ref{tab:AblationStudyMPDDVisA}. We observe that our approach, MultiADS, performs quite well even when we utilize the defect-aware text prompts of the other dataset for all the metrics on pixel-level and image-level on few-shot anomaly detection and segmentation tasks. Also, we note that to achieve the best performance, especially on the image level, it is crucial to employ defect-aware text prompts suitable for the products of the testing dataset, the MPDD dataset.

In addition to the results shown in the main manuscript, in Table~\ref{tab:multi-seg_tabels} we list the segmentation performance for some sample defect types that are seen/unseen during the training phase. 
We notice that defects such as \emph{stains} and \emph{scratches} are easy to locate and classify, as they also occur on the training dataset - MVTec-AD.  
For unseen defects like \emph{burrs} and \emph{mismatch}, our model achieves slightly lower accuracy. On the other hand, for other unseen defects such as \emph{flattening}, we perform with high precision for the classification task. These results, similar to results in the main manuscript, reflect that our approach, MultiADS, has generalization ability on large and complex datasets and unseen defects in the training dataset.

\begin{table}[t]
    \centering
    \caption{Results MTAS for zero-shot setting at pixel-level for sample defect-types. The model is trained on the MVTec-AD dataset. - indicates \textbf{unseen} defect types while \cmark indicates \textbf{seen} defect types during training.}
    \begin{subtable}[t]{0.4\textwidth}
        \centering
        \caption{MAD-sim}
        \begin{adjustbox}{width=0.99\textwidth}
        \begin{tabular}{p{0.3cm} c|c c c}
            \toprule
             & Defects & AUROC & F1-Score & AP  \\
            \hline
             - & Burrs & 95.56 & 1.18 & 1.67  \\
             \cmark & Missing & 86.52 & 2.56 & 3.08  \\
             \cmark & Stains & 98.19 & 15.02 & 9.92  \\
            \bottomrule
        \end{tabular}
        \end{adjustbox}
    \end{subtable}
    \hfill
        \begin{subtable}[t]{0.4\textwidth}
        \centering
        \caption{MPDD}
        \begin{adjustbox}{width=0.99\textwidth}
        \begin{tabular}{p{0.3cm} c| c c c}
            \toprule
             & Defects & AUROC & F1-Score & AP  \\
            \hline
             - & Mismatch & 88.44 & 2.56 & 1.04  \\
             - & Flattening & 96.72 & 36.06 & 8.33  \\
             \cmark & Scratch & 96.67 & 26.99 & 20.26  \\
            \bottomrule
        \end{tabular}
        \end{adjustbox}
    \end{subtable}
    
    \label{tab:multi-seg_tabels2}
\end{table}

\begin{table}[ht]
\centering
\caption{Image level results for batched zero-shot setting. All results are AUROC values (\%). The numbers of baselines are taken from AnomalyDINO~\cite{AnomalyDINO}. 448 and 672 are the resolutions of the input image.}
\begin{adjustbox}{width=0.4\textwidth}
\begin{tabular}{c|lcc}
\toprule
Setting & Method & MVTec & VisA \\
\midrule
\multirow{5}{*}{\makecell{Batched \\ zero-shot}} & ACR~\cite{ACR} & 85.8 & / \\
 & MuSc~\cite{MuSc} & \textbf{97.8} & 92.8 \\
 & AnomalyDINO\textsubscript{(448)}~\cite{AnomalyDINO} & 93.0 & 89.7 \\
 & AnomalyDINO\textsubscript{(672)}~\cite{AnomalyDINO} & 94.2 & 90.7 \\
\cline{2-4}
 & MultiADS (ours) & 96.1 & \textbf{93.1} \\
\bottomrule
\end{tabular}
\end{adjustbox}
\label{tab:batched zero-shot}
\end{table}

\begin{table*}[ht]
\begin{center}
\caption{Ablation study for training and testing with different architectures/resolutions for BADS. MultiADS applies the ViT-L-14 architecture with a resolution of 336.}
\label{table:ablation_study_model_resolution}
\small
\begin{tabular}{ c|c|c|c| c c c} 
\hline
Settings & \multirow{2}{*}{Dataset} & \multirow{2}{*}{Architecture} & \multirow{2}{*}{Resolution} & \multicolumn{3}{c}{Image-Level} \\
\cline{5-7}
& & & & AUROC & F1-max  & AP \\	
\hline
\multirow{8}{*}{Zero-shot} & \multirow{4}{*}{VisA} & ViT-B-16 & 224 & 74 & 76.6 & 79  \\
& & ViT-B-32 & 224 & 68.4 & 74.6 & 73.5  \\
& & ViT-L-14 & 224 & 75.2 &  78.4 &  80.6   \\
& & \cellcolor{gray!40} ViT-L-14 & \cellcolor{gray!40} 336 & \cellcolor{gray!40} 83.6  & \cellcolor{gray!40} 80.3  & \cellcolor{gray!40} 86.9  \\
\cline{2-7}
& \multirow{4}{*}{MPDD} & ViT-B-16 & 224 & 67.7 & 77.2 & 74.4 \\
& & ViT-B-32 & 224 & 60.7 & 75  &  68.8  \\
& & ViT-L-14 & 224 & 71.6 &  77.8 &  76.8  \\
& & \cellcolor{gray!40} ViT-L-14 & \cellcolor{gray!40} 336 & \cellcolor{gray!40} 78.3 & \cellcolor{gray!40} 79.2 & \cellcolor{gray!40} 78.4  \\
\hline
\multirow{8}{*}{Few-shot (k=4)} & \multirow{4}{*}{VisA} & ViT-B-16 & 224 & 90 & 86  & 91.9  \\
& & ViT-B-32 & 224 &  83.1 & 81.4 & 85.4  \\
& & ViT-L-14 & 224 & 92   & 88   & 93.5  \\
\cline{3-7}
& & \cellcolor{gray!40} ViT-L-14 & \cellcolor{gray!40} 336 &  \cellcolor{gray!40} 93.3  & \cellcolor{gray!40} 89.7  & \cellcolor{gray!40} 94.3 \\
& \multirow{4}{*}{MPDD} & ViT-B-16 & 224 & 80.2 & 81.6 & 80  \\
& & ViT-B-32 & 224 & 78.2 & 83.1 & 80.2  \\
& & ViT-L-14 & 224 & 82   &   82.9 &    84.3  \\
\cline{3-7}
& & \cellcolor{gray!40} ViT-L-14 & \cellcolor{gray!40} 336 & \cellcolor{gray!40} 85.6 & \cellcolor{gray!40} 87.2  & \cellcolor{gray!40} 89.4  \\
\hline
\end{tabular}
\end{center}
\end{table*}

\subsubsection{Batched Zero-shot Setting}

The idea behind the batched zero-shot setting is to utilize all text samples in $X_{\text{test}}$ without relying on any labels. This approach can be viewed as a form of domain adaptation, enabling the trained model to better align with the target domain. Inspired by the methodology proposed by AnomalyDINO~\cite{AnomalyDINO}, we employ a memory bank to facilitate this adaptation process.
For each test sample $x^{(k)} \in X_\text{test}$, let ${\mathbf{Z}^k_i} \in \mathbb{R}^{h\times w \times N_z}$ denote the adapted image patch embeddings at state $i$ for given image $x^{(k)}$. We define memory bank $\mathcal{M}_i$ as the union of all image patch embeddings at stage $i$ across the entire text set $X_\text{test}$:
\begin{equation}
    \mathcal{M}_i = \bigcup_{x^{(k)}\in X_\text{test}} \left\{ \mathbf{Z}^k_i[a,b] | a \in [h], b \in [w] \right\} \text{.}
\end{equation}
During testing, for each given image $x^{(k)}$, we compute the cosine similarity between its adapted image patch embedding $\mathbf{Z}^k_i[a,b] \in \mathbb{R}^{N_z}$ and all embeddings in the memory bank $\mathcal{M}_i \setminus \mathbf{Z}^k_i[a,b]$. Since the memory bank may include anomalous features (due to the unlabeled setting), directly selecting the nearest neighbor might not reliably represent nominal behavior. To address this, and based on the assumption that most patches in the memory bank are nominal, we replace the nearest neighbor with the k-th nearest neighbor, where k corresponds to the $\alpha$-quantile of the similarity scores. Thus, the set of cosine similarity scores is defined as follows:

\begin{align}
\mathcal{D}\left(\mathbf{Z}_i^k[a,b],\, \mathcal{M}_i \setminus \{\mathbf{Z}_i^k[a,b]\}\right) 
&= \left\{ d\left(\mathbf{Z}_i^k[a,b], \mathbf{x}\right) \mid \right. \nonumber\\
& \left. \mathbf{x} \in \mathcal{M}_i \setminus \{\mathbf{Z}_i^k[a,b]\} \right\}.
\end{align}
where $d(\cdot)$ represents the cosine similarity. The reference anomaly score for image patch embedding $\mathbf{Z}^k_i[a,b]$ is defined as follows:
\begin{equation}
    s(\mathbf{Z}^k_i[a,b]) = q_\alpha(\mathcal{D}(\mathbf{Z}^k_i[a,b], \mathcal{M}_i \setminus \mathbf{Z}^k_i[a,b])) \text{,}
\end{equation}
where $q_\alpha$ is the $\alpha$ quantile of the similarity score set. The comparison of our MultiADS approach with other baselines is listed in Table~\ref{tab:batched zero-shot}. 

\subsubsection{Backbones}
In Table~\ref{table:ablation_study_model_resolution}, we show the impact of different architectures and resolutions for our proposed approach, MultiADS.
To evaluate the performance of our proposed approach, MultiADS, and other baselines, we perform zero-shot and few-shot anomaly detection and segmentation on five datasets, MVTec-AD~\cite{MVTec}, VisA~\cite{VisA}, MPDD~\cite{MPDD}, MAD~\cite{PAD}, and Real-IAD~\cite{realIAD}. Results of other baselines are taken from the original published papers or the most recent publications. Thus, for some of the baselines, we are missing the evaluation with different metrics, such as F1-max, AP, and AUPRO on pixel-level, or  F1-max and AP for image-level. 

\subsubsection{Additional Results}

In Tables \ref{table:resFewVisA}, \ref{table:resFewMPDD}, and \ref{table:resFewMVTecAD}, we show results for our approach, MultiADS, and other baselines on a few-shot setting with $k \in [1, 2, 4, 8]$ on anomaly detection and segmentation tasks on three datasets, VisA, MPDD, and MVTec-AD, respectively. In Tables \ref{table:resFewVisAProduct}, \ref{table:resFewMPDDProduct}, and \ref{table:resFewMVTecADProduct}, we show results for our approach, MultiADS, on a few-shot setting with $k \in \{1, 2\}$ on anomaly detection and segmentation tasks for each product of the VisA, MPDD, and MVTec-AD datasets, respectively. In Tables \ref{table:resFewVisAProductF} and \ref{table:resFewMPDDProductF}, we show results for the variant of our approach, MultiADS-F, on the few-shot setting with $k \in \{1, 2\}$ on anomaly detection and segmentation tasks for each product of the VisA and MPDD datasets, respectively. 

Furthermore, in Table~\ref{table:resFewRealIADProduct}, we show results for our proposal, MultiADS, and the most recent baseline, AdaCLIP, for all products of the Real-IAD dataset. We note that our proposal outperforms AdaCLIP for all metrics, and the largest improvement of our method is at the image level.
Similarly, in Table \ref{table:resFewMADsimProductF}, we show results for our proposal, MultiADS, and the most competitive baseline, April-GAN, for all products of the MAD dataset. We note that our proposal overall outperforms April-GAN for almost all metrics, and the largest improvement of our method is at the pixel level.

    

\begin{table*}[t]
\begin{center}
\caption{Few-shot anomaly detection and segmentation on the VisA Datasets. April-GAN baseline and our model are trained on the MVTec-AD dataset. (- denotes the results for this metric are not reported in the original paper; \textbf{bold} represents the best performer)}\label{table:resFewVisA}
\small
\begin{adjustbox}{width=0.9\textwidth}
\begin{tabular}{ c|c| c c|c c c| c c|c c c} 
\hline
\multicolumn{2}{c|}{Settings} & \multicolumn{5}{c|}{\textbf{k=1}} & \multicolumn{5}{c}{\textbf{k=2}} \\
\hline
\multicolumn{2}{c|}{\textbf{VisA}}  & \multicolumn{2}{c|}{Pixel-Level} & \multicolumn{3}{c|}{Image-Level} & \multicolumn{2}{c|}{Pixel-Level} & \multicolumn{3}{c}{Image-Level} \\ 
\hline
 Method  & Venue  & AUROC   & AUPRO & AUROC & F1-max  & AP & AUROC   & AUPRO & AUROC & F1-max  & AP \\
 \hline		
PaDiM & \small{ICPR21} & 89.9   & 64.3 & 62.8 & 75.3 & 68.3 & 92.0   & 70.1 & 67.4 & 75.7 & 71.6\\
CoOp & \small{IJCV22} & -   & - & - & - & -  & -   & - & 83.5 & - & - \\
PatchCore & \small{CVPR23} & 95.4   & 80.5 & 79.9 & 81.7 & 82.8 & 96.1 & 82.6 & 81.6 & 82.5 & 84.8  \\
WinCLIP & \small{CVPR23} & 96.4 & 85.1 & 83.8 & 83.1 & 85.1 & 96.8  & 86.2 & 84.6 & 83.0 & 85.8 \\
April-GAN & \small{CVPR23} & 96.0 & 90.0 & 91.2 & 86.9 & 93.3 & 96.2 & 90.1 & 92.2 & 87.7 & 94.2 \\
\hline
\end{tabular}
\end{adjustbox}
\end{center}
\end{table*}

\begin{table*}[ht]
\begin{center}
\caption{Few-shot anomaly detection and segmentation on the MPDD Dataset. April-GAN baseline and our model are trained on the MVTec-AD dataset. (- denotes the results for this metric are not reported in the original paper; \textbf{bold} represents the best performer)}\label{table:resFewMPDD}
\small
\begin{adjustbox}{width=0.9\textwidth}
\begin{tabular}{ c|c| c c|c c c| c c|c c c} 
\hline
\multicolumn{2}{c|}{Settings} & \multicolumn{5}{c|}{\textbf{k=1}} & \multicolumn{5}{c}{\textbf{k=2}} \\
\hline
\multicolumn{2}{c|}{\textbf{MPDD}}  & \multicolumn{2}{c|}{Pixel-Level} & \multicolumn{3}{c|}{Image-Level} & \multicolumn{2}{c|}{Pixel-Level} & \multicolumn{3}{c}{Image-Level} \\ 
\hline
 Method  & Venue  & AUROC   & AUPRO & AUROC & F1-max  & AP & AUROC   & AUPRO & AUROC & F1-max  & AP \\
 \hline	
PaDiM & \small{ICPR21} & 73.9  & - & 57.5 & - & - & 75.4  & - & 58.0 & - & -\\
RegAD & \small{ECCV22} & 92.6  & - & 60.9 & - & - & 93.2  & - & 63.4 & - & -\\
PatchCore & \small{CVPR22} & 79.4   & -  & 68.9 & 77.2 & - & 84.4   & -  & 75.5 & 81.7 & -\\
April-GAN & \small{CVPR23} & 96.9 & 91.4 & 84.6 & \textbf{86.8} &\textbf{88.6} & 96.9 & 91.4 & 84.6 & \textbf{86.8} & 88.6 \\
GraphCore & \small{ICLR23} & 95.2 & - & \textbf{84.7} & - & -& 95.4 & - & 85.4 & - & -\\
\hline
\end{tabular}
\end{adjustbox}
\end{center}
\end{table*}

\begin{table*}[t]
\begin{center}
\caption{Few-shot anomaly detection and segmentation on the MVTec-AD Dataset. April-GAN baseline and our model are trained on the VisA dataset. (- denotes the results for this metric are not reported in the original paper; \textbf{bold} represents the best performer)}\label{table:resFewMVTecAD}
\small
\begin{adjustbox}{width=1\textwidth}
\begin{tabular}{ c|c| c c|c c c| c c|c c c} 
\hline
\multicolumn{2}{c|}{Settings} & \multicolumn{5}{c|}{\textbf{k=1}} & \multicolumn{5}{c}{\textbf{k=2}} \\
\hline
\multicolumn{2}{c|}{\textbf{MVTec-AD}}  & \multicolumn{2}{c|}{Pixel-Level} & \multicolumn{3}{c|}{Image-Level} & \multicolumn{2}{c|}{Pixel-Level} & \multicolumn{3}{c}{Image-Level} \\ 
\hline
 Method  & Venue  & AUROC   & AUPRO & AUROC & F1-max  & AP & AUROC   & AUPRO & AUROC & F1-max  & AP \\
 \hline		
PaDiM & \small{ICPR21} & 89.9   & 64.3 & 62.8 & 75.3 & 68.3 & 92.0   & 70.1 & 67.4 & 75.7 & 71.6\\
PatchCore & \small{CVPR23} & 95.4   & 80.5 & 79.9 & 81.7 & 82.8 & 96.1 & 82.6 & 81.6 & 82.5 & 84.8  \\
WinCLIP & \small{CVPR23} & \textbf{96.4} & 85.1 & 83.8 & 83.1 & 85.1 & \textbf{96.8}   & 86.2 & 84.6 & 83.0 & 85.8 \\
April-GAN & \small{CVPR23} & 96.0 & 90.0 & 91.2 & 86.9 & 93.3 & 96.2 & 90.1 & 92.2 & 87.7 & 94.2 \\
PromptAD & \small{CVPR24} & 96.7   & - & 86.9 & - & -  & 97.1   & - & 88.3 & - & - \\
\hline
\end{tabular}
\end{adjustbox}
\end{center}
\end{table*}

\begin{table*}[t]
\begin{center}
\caption{Results for MultiADS for each product of the VisA dataset on few-shot anomaly detection and segmentation tasks. Our model is trained on the MVTec-AD dataset. }\label{table:resFewVisAProduct}
\small
\begin{adjustbox}{width=1\textwidth}
\begin{tabular}{ c| c c c c|c c c| c c c c|c c c} 
\hline
Settings & \multicolumn{7}{c|}{k=1} & \multicolumn{7}{c}{k=2}\\
\hline
\textbf{VisA}  & \multicolumn{4}{c|}{Pixel-Level} & \multicolumn{3}{c|}{Image-Level} & \multicolumn{4}{c|}{Pixel-Level} & \multicolumn{3}{c}{Image-Level} \\ 
\hline
 Product    & AUROC & F1-max & AP  & AUPRO & AUROC & F1-max  & AP & AUROC & F1-max & AP   & AUPRO & AUROC & F1-max  & AP \\
 \hline		
Candle     &       98.7 &    39.7 &    25.2 &    97   &       91.2 &    88.1 &    90.8 &	98.7 &    39.3 &    24.7 &    97.1 &       92   &    88.8 &    91   \\
Capsules   &       98.1 &    47.1 &    39.9 &    90.7 &       95.4 &    92.1 &    97.6 &	98.3 &    48.8 &    44.2 &    92.9 &       96.5 &    92.5 &    98.1 \\
Cashew     &       94.6 &    49.3 &    41.8 &    96.3 &       91   &    89.7 &    95.5 &	94.3 &    49.5 &    41.4 &    96.5 &       95   &    92.2 &    97.6 \\
Chewinggum &       99.7 &    72.4 &    76.1 &    95.1 &       98.4 &    97   &    99.4 &	99.6 &    71.1 &    73.6 &    94.7 &       98.4 &    96.4 &    99.3 \\
Fryum      &       95   &    35.4 &    29.8 &    93   &       96.6 &    92.9 &    98.3 &	95.1 &    36.7 &    30.7 &    93.3 &       97.3 &    95.9 &    98.9 \\
\hline
\end{tabular}
\end{adjustbox}
\end{center}
\end{table*}

\begin{table*}[t]
\begin{center}
\caption{Results for MultiADS for each product of the MPDD dataset on few-shot anomaly detection and segmentation tasks. Our model is trained on the MVTec-AD dataset.}\label{table:resFewMPDDProduct}
\small
\begin{adjustbox}{width=1\textwidth}
\begin{tabular}{ c| c c c c|c c c| c c c c|c c c} 
\hline
Settings & \multicolumn{7}{c|}{k=1} & \multicolumn{7}{c}{k=2} \\
\hline
\textbf{MPDD}  & \multicolumn{4}{c|}{Pixel-Level} & \multicolumn{3}{c|}{Image-Level} & \multicolumn{4}{c|}{Pixel-Level} & \multicolumn{3}{c}{Image-Level} \\ 
\hline
 Product    & AUROC  & F1-max & AP & AUPRO & AUROC & F1-max  & AP & AUROC  & F1-max & AP & AUPRO & AUROC & F1-max  & AP \\
 \hline		
Bracket\_black  & 97.2 &    18.7 &    11.8 &    89.5 &       74.6 &    81.6 &    80.8 & 	98.3 &    35   &    25.3 &    94.3 &       82.4 &    82.1 &    88.9 \\
Bracket\_brown  & 96.2 &    17.6 &     8.7 &    91.1 &       53.3 &    79.7 &    71.4 &	96.2 &    19.9 &    11.1 &    90.1 &       65.8 &    81   &    78.1 \\
Bracket\_white  & 99.7 &    24.5 &    15.2 &    96.7 &       81.1 &    78.3 &    82.5 &	99.6 &    23.7 &    14.1 &    96.2 &       84.1 &    81.1 &    85   \\
Connector       & 96.4 &    33.9 &    32.4 &    87.8 &       91.4 &    82.8 &    89.3 &	96.2 &    35.1 &    34.3 &    87.7 &       93.8 &    85.7 &    91   \\
Metal\_plate    & 96.3 &    73.1 &    74.8 &    89.8 &       92   &    90.1 &    97.2 &	96.8 &    75   &    77.8 &    90.7 &       95.7 &    93.7 &    98.5 \\
\hline
\end{tabular}
\end{adjustbox}
\end{center}
\end{table*}

\begin{table*}[t]
\begin{center}
\caption{Results for MultiADS for each product of the MVTec-AD dataset on few-shot anomaly detection and segmentation tasks. Our model is trained on the VisA dataset.}\label{table:resFewMVTecADProduct}
\small
\begin{adjustbox}{width=1\textwidth}
\begin{tabular}{ c| c c c c|c c c| c c c c|c c c} 
\hline
Settings & \multicolumn{7}{c|}{k=1} & \multicolumn{7}{c}{k=2} \\
\hline
\textbf{MVTec-AD}  & \multicolumn{4}{c|}{Pixel-Level} & \multicolumn{3}{c|}{Image-Level} & \multicolumn{4}{c|}{Pixel-Level} & \multicolumn{3}{c}{Image-Level} \\ 
\hline
 Product    & AUROC  & F1-max & AP & AUPRO & AUROC & F1-max  & AP & AUROC  & F1-max & AP & AUPRO & AUROC & F1-max  & AP \\
 \hline		
Bottle     &       93.3 &    63.2 &    66.9 &    89.3 & 97.2 &    96.7 &    99.2 &	93.4 &    63.6 &    67.3 &    89.3 & 96.9 &    96.7 &    99.1 \\
Cable      & 84.8 &    37.3 &    34.1 &    81   & 82.7 &    80.8 &    90.3 &	83.8 &    39.8 &    35.1 &    80.6 & 84.6 &    82.2 &    91   \\
Capsule    & 95.3 &    36.6 &    31.1 &    93.6 & 73.6 &    93.4 &    91.6 &	95.4 &    36.7 &    30.6 &    94   & 72.9 &    93   &    91.4 \\
Carpet     & 99.1 &    73.1 &    78   &    97.3 & 99.7 &    98.3 &    99.9 &	99.1 &    72.9 &    77.6 &    97.6 & 99.8 &    98.9 &    99.9 \\
Grid       & 98.3 &    45.3 &    40.7 &    94.5 & 95.8 &    96.5 &    98.1 &	98.6 &    45.6 &    42.6 &    95.1 & 97.7 &    97.4 &    98.9 \\
\hline
\end{tabular}
\end{adjustbox}
\end{center}
\end{table*}

\begin{table*}[t]
\begin{center}
\caption{Results for MultiADS-F for each product of the VisA dataset on few-shot anomaly detection and segmentation tasks. Our model is trained on the MVTec-AD dataset. }\label{table:resFewVisAProductF}
\small
\begin{adjustbox}{width=1\textwidth}
\begin{tabular}{ c| c c c c|c c c| c c c c|c c c} 
\hline
Settings & \multicolumn{7}{c|}{k=1} & \multicolumn{7}{c}{k=2}\\
\hline
\textbf{VisA}  & \multicolumn{4}{c|}{Pixel-Level} & \multicolumn{3}{c|}{Image-Level} & \multicolumn{4}{c|}{Pixel-Level} & \multicolumn{3}{c}{Image-Level} \\ 
\hline
 Product    & AUROC & F1-max & AP  & AUPRO & AUROC & F1-max  & AP & AUROC & F1-max & AP   & AUPRO & AUROC & F1-max  & AP \\
 \hline		
Candle     & 98.7 & 40.4 & 27.1 & 97.1 & 90.4 & 84.4 & 91   & 	98.7 & 40   & 26.7 & 97   & 90.6 & 85.7 & 91.1 \\
Capsules   & 97.6 & 47.2 & 40.6 & 88.1 & 93.1 & 91.1 & 96.6 & 	97.7 & 48.2 & 42.3 & 89.6 & 93.8 & 89.7 & 96.8 \\
Cashew     & 94.1 & 39.4 & 32.1 & 96.6 & 91.7 & 89.2 & 95.7 & 	93.9 & 39.9 & 31.6 & 96.6 & 94.3 & 91.3 & 97.3 \\
Chewinggum & 99.6 & 77.6 & 82.2 & 93.1 & 98.9 & 97.4 & 99.5 & 	99.6 & 77.4 & 81.9 & 93.1 & 98.3 & 97.4 & 99.3 \\
Fryum      & 94.3 & 33.3 & 27   & 92   & 93.8 & 93.3 & 97.4 & 	94.4 & 34.1 & 27.5 & 92.3 & 94.7 & 93.8 & 98   \\
Macaroni1  & 99.5 & 35.7 & 26   & 96.2 & 89.1 & 82.4 & 91.7 & 	99.5 & 35   & 24.5 & 96.4 & 90.3 & 82.4 & 92.5 \\
Macaroni2  & 98.8 & 26.8 & 14.3 & 89.8 & 84.3 & 77.9 & 88.7 & 	98.8 & 25.5 & 13.7 & 89.3 & 82.8 & 77.2 & 86.3 \\
Pcb1       & 95.2 & 23.2 & 17.3 & 92   & 95.8 & 89.3 & 96.2 & 	95.7 & 25   & 19.1 & 92.3 & 94.9 & 87.1 & 95.4 \\
Pcb2       & 94.4 & 31   & 21.6 & 82.3 & 83.7 & 78.8 & 85.7 & 	94.5 & 35   & 24.4 & 83.3 & 87.9 & 80.4 & 90.2 \\
Pcb3       & 93.5 & 39.9 & 29.9 & 83.6 & 86.1 & 80.4 & 88   & 	93.7 & 46.1 & 35.5 & 84   & 89.6 & 83   & 90.5 \\
Pcb4       & 96.5 & 39.7 & 35.1 & 91.6 & 97.5 & 94.1 & 96.7 & 	96.5 & 40.5 & 35.4 & 91.6 & 97.4 & 94.2 & 96.5 \\
Pipe\_fryum & 97.4 & 43.4 & 34.3 & 97.7 & 99.1 & 99   & 99.4 & 	97.4 & 43   & 33.9 & 97.6 & 99   & 99.5 & 99.3 \\
\hline
Average      & 96.6 & 39.8 & 32.3 & 91.7 & 92   & 88.1 & 93.9 & 	96.7 & 40.8 & 33   & 91.9 & 92.8 & 88.5 & 94.4 \\
\hline
\end{tabular}
\end{adjustbox}
\end{center}
\end{table*}

\begin{table*}[t]
\begin{center}
\caption{Results for MultiADS-F for each product of the MPDD dataset on few-shot anomaly detection and segmentation tasks. Our model is trained on the MVTec-AD dataset.}\label{table:resFewMPDDProductF}
\small
\begin{adjustbox}{width=1\textwidth}
\begin{tabular}{ c| c c c c|c c c| c c c c|c c c} 
\hline
Settings & \multicolumn{7}{c|}{k=1} & \multicolumn{7}{c}{k=2} \\
\hline
\textbf{MPDD}  & \multicolumn{4}{c|}{Pixel-Level} & \multicolumn{3}{c|}{Image-Level} & \multicolumn{4}{c|}{Pixel-Level} & \multicolumn{3}{c}{Image-Level} \\ 
\hline
 Product    & AUROC  & F1-max & AP & AUPRO & AUROC & F1-max  & AP & AUROC  & F1-max & AP & AUPRO & AUROC & F1-max  & AP \\
 \hline		
Bracket\_black &       97.6 &    25   &    18.2 &    91.8 & 73.1 &    77.1 &    82.8 & 	98.1 &    32.1 &    23.7 &    94.1 & 78.6 &    81.1 &    86.2 \\
Bracket\_brown & 95.9 &    18.5 &     9.8 &    88.9 & 54.6 &    79.7 &    74.4 & 	95.9 &    21.1 &    13.4 &    87.9 & 65.4 &    81   &    80.6 \\
Bracket\_white & 99.6 &    22.2 &    14.1 &    95.8 & 74.6 &    78.9 &    69.8 & 	99.6 &    22.4 &    12.8 &    95.4 & 75.4 &    81.1 &    70.4 \\
Connector     & 96.3 &    30.8 &    27.3 &    87.3 & 84.8 &    70.6 &    79.8 & 	96   &    31.8 &    28.6 &    86.9 & 89   &    82.8 &    86.7 \\
Metal\_plate   & 97.6 &    80.4 &    78.3 &    93.2 & 98.4 &    97.3 &    99.4 & 	98.1 &    82.5 &    81.4 &    94.2 & 98.9 &    97.3 &    99.6 \\
\hline
\end{tabular}
\end{adjustbox}
\end{center}
\vspace{15em}
\end{table*}

\begin{table*}[t]
\begin{center}
\caption{Results for MultiADS and the most recent baseline approach, AdaCLIP, for each product of the Real-IAD dataset on few-shot (k=4) anomaly detection and segmentation tasks. Both models are trained on the MVTec-AD dataset. }\label{table:resFewRealIADProduct}
\small
\begin{adjustbox}{width=1\textwidth}
\begin{tabular}{ c| c c c c|c c c|c c c c|c c c} 
\hline
Baseline & \multicolumn{7}{c|}{MultiADS} & \multicolumn{7}{c}{AdaCLIP} \\
\hline
\textbf{Real-IAD}  & \multicolumn{4}{c|}{Pixel-Level} & \multicolumn{3}{c|}{Image-Level} & \multicolumn{4}{c|}{Pixel-Level} & \multicolumn{3}{c}{Image-Level} \\ 
\hline
 Product    & AUROC & F1-max & AP  & AUPRO & AUROC & F1-max  & AP & AUROC & F1-max & AP  & AUPRO & AUROC & F1-max  & AP  \\
 \hline		
Audiojack         &       98.4 &    54.6 &    49.9 &    89.3 &       75.8 &    72.8 &    77.8 	&	97.21	&	42.47	&	37.46	&	-	&	66.2	&	53.68	&	57.39	\\
Bottle Cap        &       99   &    41.5 &    34.9 &    92   &       81   &    71.5 &    81.3 	&	98.4	&	34.8	&	30.06	&	-	&	86.84	&	76.87	&	80.65	\\
Button Battery    &       97.5 &    47.7 &    46.7 &    89.3 &       72.9 &    75.4 &    82   	&	96.69	&	45.7	&	45.98	&	-	&	69.47	&	74.45	&	78.94	\\
End Cap           &       96   &    30.6 &    21.7 &    86.8 &       77.3 &    76.8 &    84.4 	&	90.59	&	17.74	&	7.89	&	-	&	60.45	&	74.85	&	67.59	\\
Eraser            &       99.8 &    62.2 &    63.8 &    98.6 &       92.2 &    86.2 &    92.5 	&	99.09	&	59.5	&	59.52	&	-	&	71.49	&	60.43	&	67.37	\\
\hline
\end{tabular}
\end{adjustbox}
\end{center}
\end{table*}

\begin{table*}[t]
\begin{center}
\caption{Results for MultiADS and the most competitive baseline approach, April-GAN, for each product of the MAD dataset on few-shot (k=4) anomaly detection and segmentation tasks. Both models are trained on the MVTec-AD dataset. }\label{table:resFewMADsimProductF}
\small
\begin{adjustbox}{width=1\textwidth}
\begin{tabular}{ c| c c c c|c c c|c c c c|c c c} 
\hline
Baseline & \multicolumn{7}{c|}{MultiADS} & \multicolumn{7}{c}{April-GAN} \\
\hline
\textbf{MAD}  & \multicolumn{4}{c|}{Pixel-Level} & \multicolumn{3}{c|}{Image-Level} & \multicolumn{4}{c|}{Pixel-Level} & \multicolumn{3}{c}{Image-Level} \\ 
\hline
 Product    & AUROC & F1-max & AP  & AUPRO & AUROC & F1-max  & AP & AUROC & F1-max & AP  & AUPRO & AUROC & F1-max  & AP  \\
 \hline		
Bear       &       91.8 &    16.9 &    11.9 &    82.9 &       71.9 &    93.7 &    94.6	&	91.2 &    13.1 &     8.5 &    79.8 &       64.1 &    93.5 &    92.5	\\
Bird       &       91.5 &     9.3 &     4.9 &    76.6 &       64.8 &    94.4 &    92.6	&	90.8 &     7.9 &     4.6 &    74.4 &       66.3 &    94.4 &    93.8	\\
Cat        &       94.4 &     8.7 &     4.9 &    86.4 &       57   &    94.5 &    92.3	&	94.1 &     9.2 &     5.6 &    84.5 &       58.4 &    94.5 &    92.6	\\
Elephant   &       72.5 &     6.7 &     3.8 &    67.4 &       72.9 &    93.9 &    95.8	&	71.5 &     6.7 &     3.7 &    65.7 &       64.6 &    93.9 &    94  	\\
Gorilla    &       93.3 &    11.8 &     5.9 &    82.2 &       52.1 &    96.2 &    92.7	&	92.3 &    10.1 &     5.7 &    77.3 &       55.4 &    96.2 &    93.9	\\
\hline
\end{tabular}
\end{adjustbox}
\end{center}
\end{table*}

\subsection{Visualizations}

\input{tikz/hazelnut}
\input{tikz/screw}
\input{tikz/leather}
\input{tikz/pipe_fryum}
\input{tikz/capsules}
\input{tikz/connector}
\input{tikz/tubes}
\input{tikz/phone_battery}
\input{tikz/sim_card_set}

In this subsection, we present additional visualizations of our anomaly segmentation results. We include eight examples of products from the MVTec-AD, VisA, and MPDD datasets: hazelnut (Figure~\ref{fig:visualization_hazelnut}), screw (Figure~\ref{fig:visualization_screw}), and leather (Figure~\ref{fig:visualization_leather}) from MVTec-AD; pipe\_fryum (Figure~\ref{fig:visualization_pipe_fryum}), and capsule (Figure~\ref{fig:visualization_capsule}) from VisA; and connector (Figure~\ref{fig:visualization_connector}) and tube (Figure~\ref{fig:visualization_tube}) from MPDD. All segmentation visualizations are performed in a few-shot ($k=4$) setting. Specifically, the models for hazelnut, screw, and leather were trained on the VisA dataset; the models for pipe\_fryum, capsule, and candle were trained on the MVTec-AD dataset; and the models for connector and tube were trained on the MVTec-AD dataset. We discuss some insights and limitations in the caption of these figures.
\clearpage

\end{document}